\documentclass[letterpaper]{article} % DO NOT CHANGE THIS
\usepackage{aaai2026}  % DO NOT CHANGE THIS
\usepackage{times}  % DO NOT CHANGE THIS
\usepackage{helvet}  % DO NOT CHANGE THIS
\usepackage{courier}  % DO NOT CHANGE THIS
\usepackage[hyphens]{url}  % DO NOT CHANGE THIS
\usepackage{graphicx} % DO NOT CHANGE THIS
\usepackage{natbib}  % DO NOT CHANGE THIS AND DO NOT ADD ANY OPTIONS TO IT
\usepackage{caption} % DO NOT CHANGE THIS AND DO NOT ADD ANY OPTIONS TO IT
\usepackage{algorithm}
\usepackage{algorithmic}
\usepackage{amsmath}
\usepackage{amssymb}
\usepackage{booktabs}
\usepackage{multirow}

\usepackage{comment}
\usepackage{tikz}
\usetikzlibrary{fit}
\usetikzlibrary{backgrounds}

\usepackage{xcolor}
\definecolor{darkblue}{rgb}{0.0, 0.0, 0.55}  % RGB for dark blue
\usetikzlibrary{calc}
\usepackage{adjustbox}
\usetikzlibrary{matrix, positioning, arrows.meta}

\usepackage{amsmath,amsthm}

\newtheorem{theorem}{Theorem}[section]

\newtheorem{definition}[theorem]{Definition}

\newtheorem{proposition}[theorem]{Proposition}

\newtheorem{lemma}[theorem]{Lemma}

\usepackage{newfloat}
\usepackage{listings}
\DeclareCaptionStyle{ruled}{labelfont=normalfont,labelsep=colon,strut=off} % DO NOT CHANGE THIS
\floatstyle{ruled}
\newfloat{listing}{tb}{lst}{}
\floatname{listing}{Listing}
\title{Symmetric Aggregation of Conformity Scores for Efficient Uncertainty Sets}
\author{
    Nabil Alami\textsuperscript{\rm 1,2}\equalcontrib,
    Jad Zakharia\textsuperscript{\rm 3}\equalcontrib,
    Souhaib Ben Taieb\textsuperscript{\rm 1,4}
}
\affiliations{
    \textsuperscript{\rm 1}Department of Statistics and Data Science, Mohamed Bin Zayed University of Artificial Intelligence (MBZUAI)\\
    \textsuperscript{\rm 2}CentraleSupélec, Université Paris-Saclay\\
    \textsuperscript{\rm 3}École des Ponts ParisTech\\
    \textsuperscript{\rm 4}Department of Computer Science, University of Mons\\

    nabil.alami@mbzuai.ac.ae, jad.zakharia@enpc.fr, souhaib.bentaieb@mbzuai.ac.ae
}

\begin{document}

\maketitle

\begin{abstract}
Access to multiple predictive models trained for the same task, whether in regression or classification, is increasingly common in many applications. Aggregating their predictive uncertainties to produce reliable and efficient uncertainty quantification is therefore a critical but still underexplored challenge, especially within the framework of conformal prediction (CP). While CP methods can generate individual prediction sets from each model, combining them into a single, more informative set remains a challenging problem. To address this, we propose SACP (Symmetric Aggregated Conformal Prediction), a novel method that aggregates nonconformity scores from multiple predictors. SACP transforms these scores into e-values and combines them using any symmetric aggregation function. This flexible design enables a robust, data-driven framework for selecting aggregation strategies that yield sharper prediction sets. We also provide theoretical insights that help justify the validity and performance of the SACP approach. Extensive experiments on diverse datasets show that SACP consistently improves efficiency and often outperforms state-of-the-art model aggregation baselines.

\end{abstract}
\begin{links}
\link{Code}{https://github.com/jadz1/SACP-SymmetricConformalAggregation}
%\link{Extended version}{https://aaai.org/example}
\end{links}
\section{Introduction}

Artificial Intelligence (AI) has transformed numerous domains, including computer vision, natural language processing, forecasting, and decision-making in high-stakes environments. Its success largely arises from the ability to learn complex patterns and extract meaningful representations from large, high-dimensional datasets. However, despite their impressive predictive performance, many AI models fail to provide reliable estimates of uncertainty \citep{Abdar_2021, wang2025aleatoricepistemicexploringuncertainty}. This shortcoming is particularly critical in high-risk applications, where erroneous predictions can lead to severe consequences. In such settings, it is essential to assess not only what a model predicts but also how confident it is in those predictions.

In recent years, conformal prediction (CP) has emerged as a powerful, distribution-free framework for constructing prediction sets with finite-sample coverage guarantees \citep{vovk2005algorithmic, angelopoulos2022gentleintroductionconformalprediction}. After training a model on the training data, CP uses a separate calibration set to compute nonconformity scores (NCS), which measure how well each sample conforms to, or deviates from, the patterns learned by the model. These calibration scores are assumed to be exchangeable with the test data. Under this assumption, CP provides exact marginal coverage guarantees without relying on strong distributional assumptions, making it particularly valuable in applications where reliability is critical.

When multiple predictive models are available for the same task, aggregating their outputs is a common strategy in machine learning to improve predictive performance, enhance robustness, and reduce variance \citep{dietterich2000ensemble, breiman1996bagging}. However, despite the success of model aggregation in traditional settings, its integration within the CP framework remains relatively underexplored. A central challenge lies in effectively combining the outputs of multiple conformal predictors to produce sharper prediction sets while preserving the exact coverage guarantees that make CP so appealing. In this context, efficiency refers to the tightness or expected size of the prediction sets, where smaller sets at the same coverage level indicate a more informative and useful predictor. This motivates two key questions: \textit{How can conformal predictions from multiple models be aggregated? And can such aggregation improve efficiency?}

Several methods have been proposed to address this challenge. One prominent line of work focuses on model selection, aiming to identify the most suitable predictors based on specific performance criteria \citep{jin2023selectionpredictionconformalpvalues, yang2024selectionaggregationconformalprediction, gasparin2024conformalonlinemodelaggregation}. Other studies have explored merging prediction sets from individual conformal predictors in a black-box manner \citep{gasparin2024merginguncertaintysetsmajority}. More recent approaches go further by leveraging the structure of the conformity scores produced by each model \citep{rivera2025conformalpredictionensemblesimproving, luo2025weightedaggregationconformityscores}. These methods aim to exploit shared patterns across predictors to improve efficiency, producing sharper prediction sets while maintaining valid coverage guarantees. However, some approaches rely on additional hyperparameters or fail to fully utilize all available data. Furthermore, the existing literature lacks a systematic comparison of these aggregation strategies, leaving their relative strengths and limitations insufficiently understood.

To address these gaps, we propose Symmetric Aggregated Conformal Prediction (SACP), a novel method that constructs a single prediction set by symmetrically aggregating normalized conformity scores from multiple predictors. SACP operates in two stages. First, it normalizes the scores using a transformation inspired by recent advances in e-value–based CP \citep{balinsky2024, gauthier2025values, gauthier2025backward}. Second, it applies an arbitrary symmetric aggregation function to combine the normalized scores. This design yields a flexible framework that supports data-driven aggregation strategies capable of preserving coverage while enhancing efficiency. The main contributions of this work are summarized as follows:
\begin{itemize}
    \item We introduce \textbf{Symmetric Aggregated Conformal Prediction (SACP)}, a novel method that symmetrically combines normalized NCS from multiple predictors to construct a single, informative prediction set.

    \item We develop a data-driven variant of SACP, supported by theoretical analysis, that adaptively selects aggregation strategies to improve efficiency while preserving exact marginal coverage.

    \item We conduct a comprehensive empirical evaluation on both regression and classification tasks, demonstrating that SACP consistently produces sharper prediction sets and outperforms existing conformal aggregation baselines.
\end{itemize}

\section{Background}
\label{background}

We consider a supervised learning setting with a dataset \(\{(X_i, Y_i)\}_{i \in \mathcal{I}} \subset \mathcal{X} \times \mathcal{Y}\) drawn i.i.d.\ from an unknown distribution \(P\), where \(\mathcal{X}\) is the input space, \(\mathcal{Y} \subset \mathbb{R}\) is the output space, and \(\mathcal{I}\) is the set of data indices. We split the index set \(\mathcal{I}\) into three disjoint subsets: \(\mathcal{I}_{\mathrm{train}}\), \(\mathcal{I}_{\mathrm{cal}}\), and \(\mathcal{I}_{\mathrm{test}}\), corresponding respectively to the training, calibration, and test data. We denote \(\mathcal{I}_{\mathrm{cal}} = \{1, \dots, n\}\), where \(n = |\mathcal{I}_{\mathrm{cal}}|\) is the calibration set size. We consider \(K\) base regression or classification predictors \(\hat{\mu}^{(1)}, \dots, \hat{\mu}^{(K)}\), each trained on \(\{(X_i, Y_i)\}_{i \in \mathcal{I}_{\mathrm{train}}}\). Given a new test pair \((X_{\mathrm{test}}, Y_{\mathrm{test}})\), assumed exchangeable with the calibration set \(\{(X_i, Y_i)\}_{1 \le i \le n}\), our goal is to construct a prediction set for the unknown label \(Y_{\mathrm{test}} \in \mathcal{Y}\) with a guaranteed miscoverage level \(\alpha \in (0, 1)\) while minimizing its expected length.

\paragraph{Conformal Prediction.} 
CP provides distribution-free prediction sets with finite-sample coverage guarantees \citep{vovk2005algorithmic, angelopoulos2022gentleintroductionconformalprediction}. In our setting, for each predictor \(k = 1, \dots, K\), we compute \emph{nonconformity scores} (NCS) on the calibration set \(\{(X_i, Y_i)\}_{1 \le i \le n}\), defined as \(s_i^{(k)} = s^{(k)}(X_i, Y_i)\), where \(s^{(k)} : \mathcal{X} \times \mathcal{Y} \to \mathbb{R}_+\) measures how atypical the true label \(Y_i\) is with respect to the model prediction \(\hat{\mu}^{(k)}(X_i)\). For regression, we set \(s^{(k)}(X, Y) = |Y - \hat{\mu}^{(k)}(X)|\), and for classification, \(s^{(k)}(X, Y) = 1 - \hat{\mu}^{(k)}_Y(X)\), where $\hat{\mu}^{(k)}(X)$ and \(\hat{\mu}^{(k)}_Y(X)\) denote the estimated regression function and the predicted probability of class \(Y\), respectively. Alternative definitions of NCS can also be used; see \citet{Dheur2025-unified} for examples. Denote by \(s^{(k)}_{(1)} \le \cdots \le s^{(k)}_{(n)}\) the order statistics of the calibration scores, and let \(\hat{Q}^k_\alpha = s^{(k)}_{(\lceil (1 - \alpha)(n + 1) \rceil)}\) be their empirical quantile, computed from \(\mathcal{S}_{\mathrm{cal}} := \{s_i^{(k)}\}_{1 \le i \le n}\). The corresponding prediction set is defined as
\begin{align}
\mathcal C^k_\alpha(X_{\mathrm{test}}) &= \{\, y \in \mathcal{Y} \mid s^{(k)}(X_{\mathrm{test}}, y) \le \hat{Q}^k_\alpha \}, \label{eq:pred_nonconf}
\end{align}
which guarantees finite-sample marginal coverage:
\begin{equation}
\label{Valid_cov}
\mathbb{P}\big(Y_{\mathrm{test}} \in \mathcal C^k_\alpha(X_{\mathrm{test}})\big) \ge 1 - \alpha.
\end{equation}
Alternatively, one can define \(\tilde{s}^{(k)} = -s^{(k)}\) and express the prediction set as \(\mathcal C^k_\alpha(X_{\mathrm{test}}) = \{\, y \in \mathcal{Y} \mid \tilde{s}^{(k)}(X_{\mathrm{test}}, y) \ge q^k_\alpha \}\), with \(q^k_\alpha = \tilde{s}^{(k)}_{(\lfloor (n + 1)\alpha \rfloor)}\), which still satisfies \eqref{Valid_cov}. This formulation highlights that the resulting prediction set depends on the choice of NCS.

\paragraph{CP aggregation methods.}

Aggregating predictors within the CP framework aims to combine the outputs of multiple models trained for the same task into a single unified prediction set. Since efficiency is also a key objective, aggregation methods typically aim to produce the smallest possible prediction sets while maintaining valid coverage guarantees. Broadly, these approaches can be grouped into two main classes:

%\noindent\textbf{1. Combining at the prediction set level:} 

\noindent\textbf{1. Combining prediction sets}.  A simple aggregation strategy combines the individual prediction sets either by \emph{intersection}, 
\(\bigcap_{k=1}^K \mathcal{C}_\alpha^k\), which reduces the set size but yields coverage of at least \(1 - K\alpha\), 
or by \emph{union}, \(\bigcup_{k=1}^K \mathcal{C}_\alpha^k\), which guarantees coverage at the cost of a larger prediction set. 
%\citet{yang2024selectionaggregationconformalprediction} proposed a model selection approach that chooses the single best predictor by minimizing the expected prediction set length:
%\begin{equation}\mathcal{C}^{\text{Sel}}_\alpha(X_{\text{test}}) := \mathcal{C}^{\,k^\star}_\alpha(X_{\text{test}}), \quad k^\star \in \arg\min_{1 \le k \le K} \,\mathbb{E}\!\left[\,\big|\mathcal{C}^k_\alpha(X_{\text{test}})\big|\,\right].\end{equation}

A more intuitive alternative is the \emph{majority vote} method introduced by \citet{gasparin2024merginguncertaintysetsmajority}, which includes a candidate label if it is accepted by most individual predictors:
\begin{equation}
\mathcal{C}_\alpha^M(X_{\text{test}}) := 
\left\{
  y \in \mathcal{Y} \mid
  \frac{1}{K} \sum_{k=1}^K \mathbf{1}_{[y \in \mathcal{C}_\alpha^k(X_{\text{test}})]} > \frac{1}{2}
\right\},
\end{equation}
achieving coverage of at least \(1 - 2\alpha\). 
To further improve efficiency, the authors proposed a randomized variant in which the fixed majority threshold \(1/2\) is replaced by \(1/2 + U\), with \(U \sim \text{Unif}[0, 1]\). 
While elegant, the majority vote approach operates solely on the final prediction sets, limiting its ability to fully exploit the information available at the score level.

\begin{table*}[ht]
\centering
\small 
\setlength{\tabcolsep}{4pt} 
\begin{tabular}{lccll}
\toprule
\textbf{Method} & \textbf{Coverage Guarantee} & \textbf{Time complexity} & \textbf{Calibration} & \textbf{Notes} \\
\midrule
Wagg & $1-\alpha$ & $\mathcal{O}(W_{\text{grid}} \cdot n_1 + n_2 \cdot W_{\text{grid}} \cdot D_{\text{grid}} + n_3)$ & Split in 2 or 3 & Challenging in high dimension \\
CSA & $1-\alpha$ & $\mathcal{O}(M \cdot K \cdot n)$ & Split in 2 & Based on projected quantiles \\
CM \& CR & $1-2\alpha$ & $\mathcal{O}(K \cdot n)$ & No split & Set-level symmetric aggregation \\
SACP & $1-\alpha$ & $\mathcal{O}(D_{\text{grid}} \cdot K \cdot n)$ & No split & Choice of aggregating function \\
\bottomrule
\end{tabular}

\caption{Comparison of aggregation methods. Time complexity refers to the time to generate the prediction set of a single test point.
\textbf{Notation:} $n_i$ --- calibration size of subset $i$ of the calibration set; $n$ --- total calibration size; $K$ --- number of predictors; $W_{\text{grid}}$ --- size of the grid for weights; $M$ --- number of directions. $D_{\text{grid}}$ — number of classes (in classification) or length of the discretized target space $\mathcal{Y}$ (in regression).
}
\label{tab:agg_methods}
\end{table*}

\noindent\textbf{2. Combining scores.} Another class of methods aggregates information directly from the NCS. 
\citet{luo2025weightedaggregationconformityscores} propose aggregating NCS through a convex combination. For any pair \((X, Y)\), let \(s(X, Y) = (s^{(1)}(X, Y), \ldots, s^{(K)}(X, Y))^\top\) denote the corresponding vector of NCS.  The aggregated score is then given by \(s(X, Y)^\top \mathbf{w}\), where \(\mathbf{w}\) lies on the \((K-1)\)-simplex (i.e., \(w_k \ge 0\) and \(\sum_{k=1}^K w_k = 1\)) and is chosen to minimize the expected prediction set length. The corresponding prediction set is defined as
\begin{equation}
    \mathcal{C}^{\mathbf{w}}_\alpha(X_{\text{test}}) := 
    \left\{ y \in \mathcal{Y} \;\middle|\; s(X_{\text{test}}, y)^\top \mathbf{w} \le \hat{Q}_\alpha \right\},
\end{equation}
where \(\hat{Q}_\alpha\) is the empirical \((1 - \alpha)\)-quantile of the aggregated scores, computed from a subset of the calibration data to ensure valid coverage. 

\citet{rivera2025conformalpredictionensemblesimproving} propose \emph{Conformal Score Aggregation (CSA)}, a method based on multivariate quantiles. 
The calibration set is split into two subsets, \(\mathcal{S}_{\mathrm{cal}} := \mathcal{S}^{(1)} \cup \mathcal{S}^{(2)}\). 
On \(\mathcal{S}^{(1)}\), CSA samples \(M\) directions \(\{u_m\}_{m=1}^M\) on the positive unit sphere and determines thresholds \(q_m\) such that a fraction \(\beta\) of projections \(u_m^\top s\) exceed each \(q_m\). 
These thresholds define the envelope
\[
H(\beta) = \bigcap_{m=1}^M \left\{ s \in \mathcal{S}^{(1)} : u_m^\top s \le q_m(\beta) \right\},
\]
and a binary search over \(\beta\) is performed to achieve \(1 - \alpha\) coverage on \(\mathcal{S}^{(1)}\). 
Next, each score \(s \in \mathcal{S}^{(2)}\) is mapped to \(\max_m \left( \frac{u_m^\top s}{q_m(\beta^*)} \right)\); rescaling \(H(\beta^*)\) by the empirical \((1 - \alpha)\)-quantile of these normalized values preserves exchangeability and guarantees exact coverage. 
Table~\ref{tab:agg_methods} summarizes these aggregation methods.

\paragraph{Setup Justification.} In the context of aggregation, the key components are the training data, the calibration data, and the NCS.  Differences in any of these elements can lead to distinct CP settings.  To ensure a fair and consistent comparison across methods, our setup uses \(K\) predictors trained on the same training dataset, evaluated on the same calibration set using a fixed nonconformity score.

\section{Our SACP Method}
\label{sec:SACP}

Our SACP method aims to construct a single final prediction set from \(K\) base predictors. Recent advances in CP have underscored the importance of \(e\)-values for uncertainty quantification \citep{gauthier2025values}. Building on this idea, SACP  normalizes the NCSs by transforming them into \(e\)-variables. They are then combined through a symmetric aggregation function, enabling the use of standard (split) CP.

\begin{definition}
    An e-variable $E$ is a nonnegative random variable $(E\geq0)$ that satisfies, under a null hypothesis $\mathbb{H}_0$,
\begin{equation}
    \mathbb{E}_{\mathbb H_0}[E] \leq 1.
\end{equation}
An e-value is the observed value of an e-variable.
\end{definition}

The first step of SACP is to construct e-variable–like transformations of the NCSs, inspired by the work of \citet{balinsky2024}, although we do not rely on the general e-value theory.
This construction is based on the following proposition.
\begin{proposition}
Consider exchangeable and positive random variables \(\{S_i\}_{1 \leq i \leq n+1}\). Then, the random variables 
\begin{equation}
E_j = \frac{S_j}{\frac{1}{n+1} \sum_{i=1}^{n+1} S_i},\quad  j = 1,\dots ,n
\label{evalue_Balinsky}
\end{equation}
have expectation equal to one.
\end{proposition}

Specifically, for each predictor and its corresponding scores, SACP constructs \(n+1\) exchangeable \(e\)-variables of the form~\eqref{evalue_Balinsky}.

Let \(y \in \mathcal{Y}\) denote a candidate label associated with \(X_{\mathrm{test}}\), so that \(s^{(k)}(X_{\mathrm{test}}, y)\) quantifies its conformity under predictor \(k\). 
For each \(k = 1, \dots, K\) and each calibration point \((X_i, Y_i)\), \(i = 1, \dots, n\), we construct the calibration \(e\)-variable
\begin{equation}
    E^{(k)}_i(y) = 
    \frac{s^{(k)}(X_i, Y_i)}
    {\frac{1}{n+1}\!\left(\sum_{j=1}^n s^{(k)}(X_j, Y_j) + s^{(k)}(X_{\text{test}}, y)\right)}.
\end{equation}
This quantity is the ratio of the \(i^{\text{th}}\) score from predictor \(k\) to the average of all calibration scores augmented with the test score for the candidate \(y\). 

Since the scores \(\{s^{(k)}(X_i, Y_i)\}_{1 \le i \le n}\) are exchangeable, \(E_i^{(k)}\) follows the same form as in~\eqref{evalue_Balinsky} and is therefore an \(e\)-variable (under the null hypothesis $\mathbb{H}_0:y=Y_{\text{test}}$, and assuming the NCS are positive). 
For each calibration point \((X_i, Y_i)\), we define its corresponding \(e\)-vector as 
\(\mathbf{E}_i = (E^{(1)}_i, \dots, E^{(K)}_i)^{\top} \in \mathbb{R}^K\).
Similarly, the test \(e\)-variable is defined as
\begin{equation}
    E^{(k)}_{\text{test}}(y) = 
    \frac{s^{(k)}(X_{\text{test}}, y)}
    {\frac{1}{n+1}\!\left(\sum_{i=1}^n s^{(k)}(X_i, Y_i) + s^{(k)}(X_{\text{test}}, y)\right)},
\end{equation}
and the corresponding test \(e\)-vector is 
\(\mathbf{E}_{\text{test}} = (E^{(1)}_{\text{test}}, \dots, E^{(K)}_{\text{test}})^{\top}\).

Note that all \(\{E_i^{(k)}\}_{1 \le i \le n}\) share the same denominator, so their relative ordering is preserved. 
Moreover, the test \(e\)-variable mirrors the behavior of the test score, decreasing as \(s^{(k)}(X_{\text{test}}, y)\) decreases.

The second step of SACP is to combine the constructed \(e\)-variables using a \textit{symmetric} function $f: \mathbb{R}^K \longrightarrow \mathbb{R}$, which merges the \(e\)-variables into new aggregated scores. For \(i = 1, \ldots, n\), we define
\begin{equation}
    F_i(y) := f(\mathbf{E}_i(y)), 
    \quad 
    F_{\text{test}}(y) := f(\mathbf{E}_{\text{test}}(y)). \label{eq:F}
\end{equation}

%If \(f\) is non-decreasing in each component, the resulting aggregated score behaves as a valid nonconformity measure, where lower values indicate better conformity. In this case, the prediction set is given by 

Working with symmetric aggregators is especially natural in a distribution-free setting and offers two key advantages. First, the aggregated score is invariant to how the base models are indexed or labeled: permuting their indices leaves the aggregated value unchanged, ensuring that model labeling has no effect. Second, symmetry confines us to a structured and interpretable class of functions, which in turn makes it easier (a) to optimize the aggregator in practice and (b) to derive theoretical results on the resulting prediction-set size.

Note that the aggregated scores behavior directly depends on the choice of the aggregation function. When the aggregated scores exhibit the characteristics of a classical NCS, in which smaller values correspond to higher conformity, the prediction set is defined as
\begin{equation}
    \label{Predset_lower}
\mathcal{C}_\alpha^{f,\,\uparrow}(X_{\text{test}})
=\{y \in \mathcal{Y} \mid F_{\text{test}}(y)\leq \hat{Q}_\alpha(y) \},
\end{equation}
where 
\begin{equation}
\label{empirical_quantile_up}
\hat{Q}_{\alpha}(y) := F_{(\lceil (1 - \alpha)(n + 1) \rceil)}(y)
\end{equation}
is the upper empirical quantiles of $\{F_i\}_{1\le i\le n}$. In contrast, when higher values indicate better conformity, the prediction set is
\begin{equation}
     \label{Predset_greater}
\mathcal{C}_\alpha^{f,\,\downarrow}(X_{\text{test}})
=\{y \in \mathcal{Y} \mid F_{\text{test}}(y)\geq \hat{q}_\alpha(y) \},
\end{equation}
where
\begin{equation}
\label{empirical_quantile_low}
\hat{q}_{\alpha}(y) := F_{(\lfloor \alpha(n + 1) \rfloor)}(y)
\end{equation}
is the lower empirical quantiles of $\{F_i\}_{1\le i\le n}$.

\begin{theorem}
    \label{main_theorem}
The SACP prediction sets defined in \eqref{Predset_lower} and \eqref{Predset_greater} verify the coverage property \eqref{Valid_cov}.
\end{theorem}

\begin{figure}[h]
    \centering
    \includegraphics[width=0.95\linewidth]{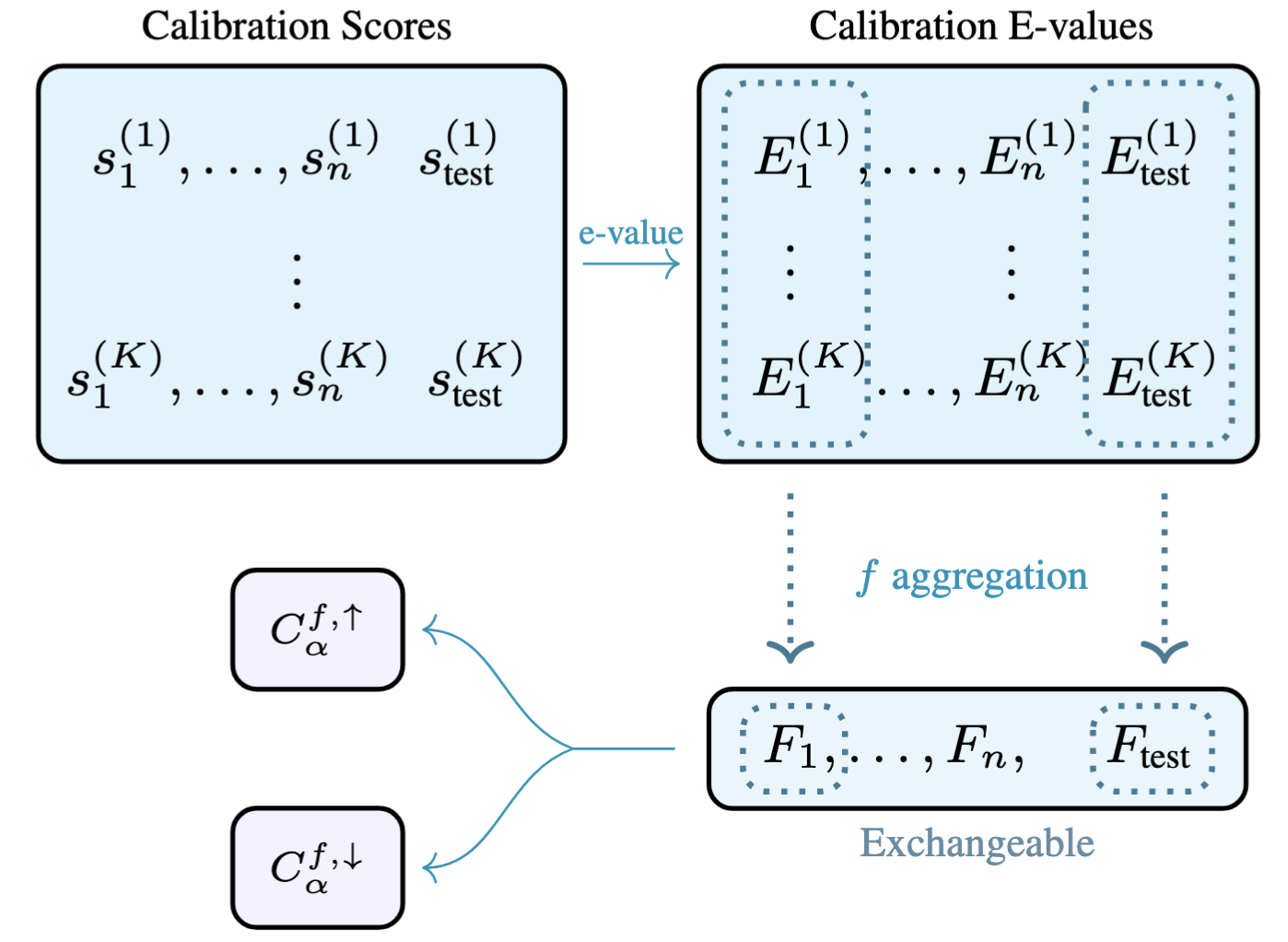}
  \caption{Diagram illustrating key steps of SACP}
\label{Overview_SACP}
\end{figure}

\noindent The proof follows from the exchangeability of \(\{F_i\}_{1 \le i \le n} \cup F_{\text{test}}\), see Appendix A. In our implementation of SACP, the default choice for \(f\) is the sum of its inputs. An overview of the main steps of SACP is presented in Figure~\ref{Overview_SACP}.

\noindent\textbf{Remark 1.} If we interpret the original NCS as “smaller values correspond to better conformity,” then indeed the behavior of the aggregated score aligns with the monotonicity of $f$. If $f$ is non-decreasing component-wise, the resulting merged score behaves as a NCS: lower values indicate better conformity. In this case, the prediction set is constructed using \eqref{Predset_lower}. In contrast, if $f$ is decreasing component-wise, the interpretation is reversed: higher values indicate better conformity, and the prediction set is built according to \eqref{Predset_greater}. If  $f$ is non-monotonic, it is difficult to draw precise conclusions about the behavior of the aggregated scores. 

\noindent\textbf{Remark 2.}  Unlike the standard CP framework, the thresholds \(\hat{q}_{\alpha}\) and \(\hat{Q}_{\alpha}\) depend on the candidate label \(y\). In classification tasks, a separate quantile must be computed for each class, whereas in regression we discretize the output space \(\mathcal{Y}\) into a uniform grid and compute one quantile for each grid point.

Our SACP method offers several key advantages and is particularly relevant in the context of conformal aggregation:
\begin{itemize}
    \item To the best of our knowledge, this is the first approach to apply symmetric conformal aggregation at the score level while guaranteeing \(1 - \alpha\) coverage. This is achieved without requiring any additional data splitting of the calibration set (see Table~\ref{tab:agg_methods}).

    \item Transforming raw scores into \(e\)-variables standardizes them across models by ensuring a common first moment (i.e., an expected value equal to one), enabling fair aggregation regardless of differences in scale or distribution.

    \item The method is conceptually simple and allows a flexible choice of the symmetric function, paving the way for further improvements through data-driven aggregation strategies.
\end{itemize}

\subsection{Theoretical analysis}
\label{sec:theory}

The aggregation of scores using an arbitrary symmetric function is subtle and requires careful analysis. In this section, we examine the influence of the symmetric aggregation function and explore key properties of the empirical quantile. These results allow us to derive a worst-case bound for our prediction set in regression tasks. All proofs are provided in Appendix A.%\ref{appendix:proofs}

\subsubsection{The aggregating function.}

SACP requires selecting a symmetric function; however, not all symmetric functions yield more informative prediction sets. Since analyzing the general structure of symmetric functions is inherently difficult, we follow the framework of \citet{zaheer2018deepsets} and focus on the following subclass:
\begin{equation}
\label{Deepset}
    f(\mathbf{x}) := (\rho \circ \Phi)(\mathbf{x}) = \rho\left(\sum_{k=1}^K \phi(x_k)\right),
\end{equation}
where $\mathbf{x} = (x_1, \dots, x_K)^T \in (\mathbb{R}_+)^K$, and $\phi$ and $\rho$ are continuous scalar transformations.  

As discussed earlier, the monotonicity of \( f \) plays a crucial role in determining how the aggregated scores behave like NCS. Therefore, we restrict attention to strictly monotonic functions \( \rho \) and \( \phi \).  

If \( f \) is increasing, then \( \rho \) and \( \phi \) share the same monotonicity. In this case, the aggregated scores \( F_i \) in \eqref{eq:F} behave like NCS, and the corresponding prediction set is given in \eqref{Predset_lower}. Conversely, if \( f \) is decreasing, \( \rho \) and \( \phi \) have opposite monotonicity, and the associated prediction set is given in \eqref{Predset_greater}. Monotonic transformations are particularly convenient, as their influence on the ordering of scores is simpler to analyze.

\begin{proposition}
\label{prop:V_ordering}
Let \(V_{(1)} \le \cdots \le V_{(n)}\) be the order statistics of \(V_1, \dots, V_n \in \mathbb{R}\).  
Let \(g:\mathbb{R} \to \mathbb{R}\) be continuous and define \(\tilde V_i = g(V_i)\); denote their order statistics by \(\tilde V_{(1)} \le \cdots \le \tilde V_{(n)}\). Then, for any \(\alpha \in (0,1)\),
\begin{itemize}
    \item If \( g \) is strictly decreasing, then
    \begin{equation}
        \tilde{V}_{(\lceil (n+1)(1-\alpha) \rceil)}
        \;=\;
        g\big(V_{(\lfloor (n+1)\alpha \rfloor)}\big).
    \end{equation}

    \item If \( g \) is strictly increasing, then
    \begin{equation}
        \tilde{V}_{(\lceil (n+1)(1-\alpha) \rceil)}
        \;=\;
        g\big(V_{(\lceil (n+1)(1-\alpha) \rceil)}\big).
    \end{equation}
\end{itemize}
\end{proposition}

\noindent This proposition directly implies the following result.

\begin{proposition}
\label{prop:rho_composition}
For any \(\alpha \in (0,1)\), consider a symmetric function of the form \eqref{Deepset}, where \(\rho\) and \(\phi\) are monotonic. Then, for a test point \( X_{\text{test}} \), we have:
\begin{itemize}
    \item If \(\rho\) is strictly non-decreasing:
    \begin{equation}
        \mathcal{C}_\alpha^{\rho \circ \Phi,\,\uparrow}(X_{\text{test}}) = \mathcal{C}_\alpha^{\Phi,\,\uparrow}(X_{\text{test}}),
    \end{equation}
    \begin{equation}
        \mathcal{C}_\alpha^{\rho \circ \Phi,\,\downarrow}(X_{\text{test}}) = \mathcal{C}_\alpha^{\Phi,\,\downarrow}(X_{\text{test}}).
    \end{equation}

    \item If \(\rho\) is strictly non-increasing:
    \begin{equation}
        \mathcal{C}_\alpha^{\rho \circ \Phi,\,\uparrow}(X_{\text{test}}) = \mathcal{C}_\alpha^{\Phi,\,\downarrow}(X_{\text{test}}),
    \end{equation}
    \begin{equation}
        \mathcal{C}_\alpha^{\rho \circ \Phi,\,\downarrow}(X_{\text{test}}) = \mathcal{C}_\alpha^{\Phi,\,\uparrow}(X_{\text{test}}).
    \end{equation}
\end{itemize}
\end{proposition}

\noindent Therefore, these results show that the dependence on \(\rho\) can be eliminated when constructing our prediction set: $ \mathcal{C}_\alpha^{\rho \circ \Phi} = \mathcal{C}_\alpha^{\Phi}$.

\noindent For this reason, we consider the following class of symmetric functions.

\begin{definition}
\label{def:phi_agg}
A function \(\Phi\) is called a \emph{\(\phi\)-aggregating function} if it can be expressed as
\begin{equation}
\label{Deepset_simple}
    \Phi(\mathbf{x}) := \sum_{k=1}^K \phi(x_k),
\end{equation}
for all \(\mathbf{x} = (x_1, \dots, x_K)^\top \in (\mathbb{R}_+)^K\), where \(\phi : \mathbb{R}_+ \to \mathbb{R}\) is continuous and monotonic.  
The set of all \(\phi\)-aggregating functions is denoted by \(\mathcal{F}_{\mathrm{agg}}\).
\end{definition}

\noindent From this point onward, we restrict our attention to aggregating functions of the form given in Definition~\ref{def:phi_agg}. We next establish a bound on the length of our prediction set, which requires intermediate results on the empirical quantile.

\subsubsection{Quantile aggregation}

Aggregating quantiles under arbitrary dependence is a classical problem in statistics and probability theory, with early work dating back to the early 20th century \citep{vincent1912functions, thomas1980appropriate}. More recent research has established theoretical bounds and relationships between individual quantiles and their aggregated counterparts \citep{lichtendahl2013better, blanchet2024convolution}. 

The SACP methods involves computing the empirical quantile of the aggregated scores, which we will seek to upper bound in order to establish our final theorem. 

We now present an upper bound on the length of the SACP prediction set.  
Since the length naturally depends on the choice of the function \(\phi\) in \eqref{Deepset_simple}, quantifying this dependence exactly is often intractable.  
Instead, we derive a \textbf{worst-case bound} that holds uniformly over all admissible choices of \(\phi\).  
This bound characterizes the maximal width that any aggregated prediction set can attain, ensuring that any specific choice of \(\phi\) can only improve (i.e., tighten) the prediction set relative to this limit.

\begin{theorem}[Worst-case bound]
\label{theorem : bound C^f_alpha}
For a regression setting having $K>1$ predictors, and using the absolute residual NCSs, consider \(\Phi \in \mathcal{F}_{\mathrm{agg}}\).  
For a test point \(X_{\mathrm{test}}\), denote by \(|\mathcal{C}^{\Phi}_\alpha(X_{\mathrm{test}})|\) the length of the SACP prediction set, and by \(|\mathcal{C}^k_\alpha(X_{\mathrm{test}})|\) the length of the CP set of the \(k\)-th predictor \(\hat{\mu}^{(k)}\) at level $\alpha\in[\frac{K}{n+1},1)$. Let $\alpha' :=\alpha/K,$ and  
define the model disagreement at \(X_{\mathrm{test}}\) as
\begin{equation}
\label{model_disagreement}
    \Delta_{\mathrm{test}}
    = \max_{1 \le k \le K} \hat{\mu}^{(k)}(X_{\mathrm{test}})
    - \min_{1 \le k \le K} \hat{\mu}^{(k)}(X_{\mathrm{test}}).
\end{equation}
Then,  
\begin{equation}
\label{final_bound_1}
   \big|\mathcal{C}^{\Phi}_\alpha(X_{\mathrm{test}})\big|
   \;\le\;
   \Delta_{\mathrm{test}}
   + \max_k \big|\mathcal{C}^k_{\alpha'} (X_{\mathrm{test}})\big|.
\end{equation}

\end{theorem}

\noindent
The proof of Theorem~\ref{theorem : bound C^f_alpha} relies on bounding the quantities that appear in the SACP prediction set. The inequality \eqref{final_bound_1} provides a worst-case guarantee on the length of the aggregated prediction set. 

%cannot be wider than the range of model predictions plus the largest individual conformal interval. When the models are well-aligned (i.e., \(\Delta_{\mathrm{test}}\) is small), this bound indicates that aggregation will not inflate the prediction interval and may, in practice, yield tighter sets, while still preserving the validity guarantee of CP. Nonetheless, the theorem itself provides only an upper bound on the prediction set width, not a formal guarantee of improvement.

\subsection{An efficiency-oriented SACP method}
\label{sec : enhancement}

We now propose an efficiency-oriented extension of SACP, building upon the theoretical guarantees established earlier. By Theorem~\ref{main_theorem}, the prediction set \(\mathcal{C}_\alpha^\Phi\) achieves the desired \(1-\alpha\) coverage for any symmetric aggregation function \(\Phi \in \mathcal{F}_{\mathrm{agg}}\).  
This invariance allows us to exploit the flexibility in choosing \(\Phi\) to improve efficiency, specifically, to minimize the expected length of the prediction sets.

Formally, we aim to find an aggregation function \(\Phi^*\) that minimizes the prediction set length  \begin{equation}
    \Phi^* = \arg\min_{\Phi \in \mathcal{F}_{\mathrm{agg}}} \, |\mathcal{C}_\alpha^\Phi|.
\end{equation}However, searching over the entire space \(\mathcal{F}_{\mathrm{agg}}\) is intractable.  
We therefore restrict attention to the parametric subclass
\begin{equation}
    \Phi_p(\mathbf{x}) = \sum_{k=1}^K (x_k)^p, 
    \qquad p \in \mathbb{R},
\end{equation}
which encompasses several standard aggregation schemes.  
For instance, as \(p \to +\infty\), \(\Phi_p\) approaches the maximum operator, while \(p \to -\infty\) yields the minimum.  
This family has also been successfully employed in CP for efficiency optimization \citep{braun2025minimumvolumeconformalsets}.

We then select the exponent \(p^*\) that minimizes the average prediction set length on the (unlabeled) test set:
\begin{equation}
    \label{p_opt}
    p^* = \arg\min_{p \in \mathbb{R}} 
    \frac{1}{|\mathcal{I}_\text{test}|}
    \sum_{i \in \mathcal{I}_\text{test}} 
    |\mathcal{C}_\alpha^{\Phi_p}(X_i)|.
\end{equation}

\noindent
Since Theorem~\ref{main_theorem} guarantees valid coverage for any symmetric aggregator, this optimization preserves coverage while enhancing efficiency.  
Among all coverage-preserving functions \(\Phi_p\), we thus select the exponent \(p^*\) that minimizes the average set length.  
We refer to this enhanced variant as \textbf{SACP++}.  
Recall that the default SACP corresponds to \(p = 1\).

%Thus, using the test set allows us to select the function $\Phi^*$ that minimizes the average prediction set size in practice, without compromising theoretical validity.

\section{Related Work}

Several approaches have been proposed to aggregate conformal predictions.  
As discussed in Section~\ref{background}, some methods focus on aggregating prediction sets \citep{ gasparin2024merginguncertaintysetsmajority}, while others operate directly on the \emph{scores} \citep{luo2025weightedaggregationconformityscores, rivera2025conformalpredictionensemblesimproving}. Our work is also related to conformal selection \citep{yang2024selectionaggregationconformalprediction}, which can introduce selection bias.
We next review additional contributions in this area, highlighting their respective strengths and limitations.

\medskip

A number of works in the literature define multivariate NCSs for CP.  
Although not explicitly designed for aggregation, these methods naturally extend to it by mapping scores from \(\mathbb{R}^K\) to \(\mathbb{R}\).  
A recent line of research leverages \emph{Optimal Transport} to address multivariate score spaces \citep{klein2025multivariateconformalpredictionusing, thurin2025optimaltransportbasedconformalprediction}.  
These approaches construct transport maps that project multivariate NCSs onto a single radial axis, enabling the application of standard univariate calibration procedures.  
While conceptually elegant and theoretically well-grounded, they can be computationally intensive and demand careful algorithmic design to ensure numerical stability and robust performance.

\medskip

Our method involves the construction of NCSs inspired by \emph{e-values}.  
E-values have recently gained attention in the CP community as a powerful alternative to p-values \citep{vovk2025, wang2022false, ramdas2024hypothesis, balinsky2024, gauthier2025values, gauthier2025backward}.  
Rooted in hypothesis testing, e-values enable more natural and admissible aggregation schemes \citep{Merg_e_val, wang2024only}.  
They are particularly appealing due to their flexibility, anytime validity, robustness, post-hoc guarantees, and deep connections to betting and martingale theory.  
In contrast, their traditional counterparts, p-values, have long served as the cornerstone of CP but exhibit limitations when multiple p-values must be combined under arbitrary dependence structures \citep{vovk2020combining, vovk2022admissible, gasparin2025combining, ramdas2024hypothesis}.  
While certain assumptions, such as exchangeability, can partially restore statistical power through refined inequalities \citep[e.g.,][]{gasparin2025improving}, p-values remain less flexible for aggregation in general settings.

\section{Experiments and Results}

We conduct a large-scale experimental study evaluating our methods, \textbf{SACP} and \textbf{SACP++}, on both regression and classification tasks, and compare their performance against several conformal aggregation techniques.  

\paragraph{Datasets.}
For regression, we use a benchmark of single-output regression datasets from OpenML \citep{vanschoren2014openml}, previously used in CP studies \citep{rivera2025conformalpredictionensemblesimproving}.  
For classification, we evaluate on \textbf{CIFAR-10} \citep{krizhevsky2009learning} and \textbf{MNIST} \citep{lecun2010mnist}, two standard benchmarks widely used in CP research \citep{luo2025weightedaggregationconformityscores, luo2024trustworthyclassificationrankbasedconformal}.

\paragraph{Baselines.}
We compare against four aggregation methods described in Section~\ref{background}:  
(i) the weighted aggregation approach of \citet{luo2025weightedaggregationconformityscores}, denoted \textbf{Wagg};  
(ii) the multivariate-quantile method of \citet{rivera2025conformalpredictionensemblesimproving}, denoted \textbf{CSA};  
(iii) the deterministic and randomized majority-vote merging strategies of \citet{gasparin2024merginguncertaintysetsmajority}, denoted \textbf{CM} and \textbf{CR}, respectively; and  
(iv) a best-model selection rule, denoted \textbf{BL}, which selects the average empirical coverage and set length of the model achieving the smallest prediction-set length.
\begin{figure}[H]
    \centering
    \includegraphics[width=1\columnwidth]{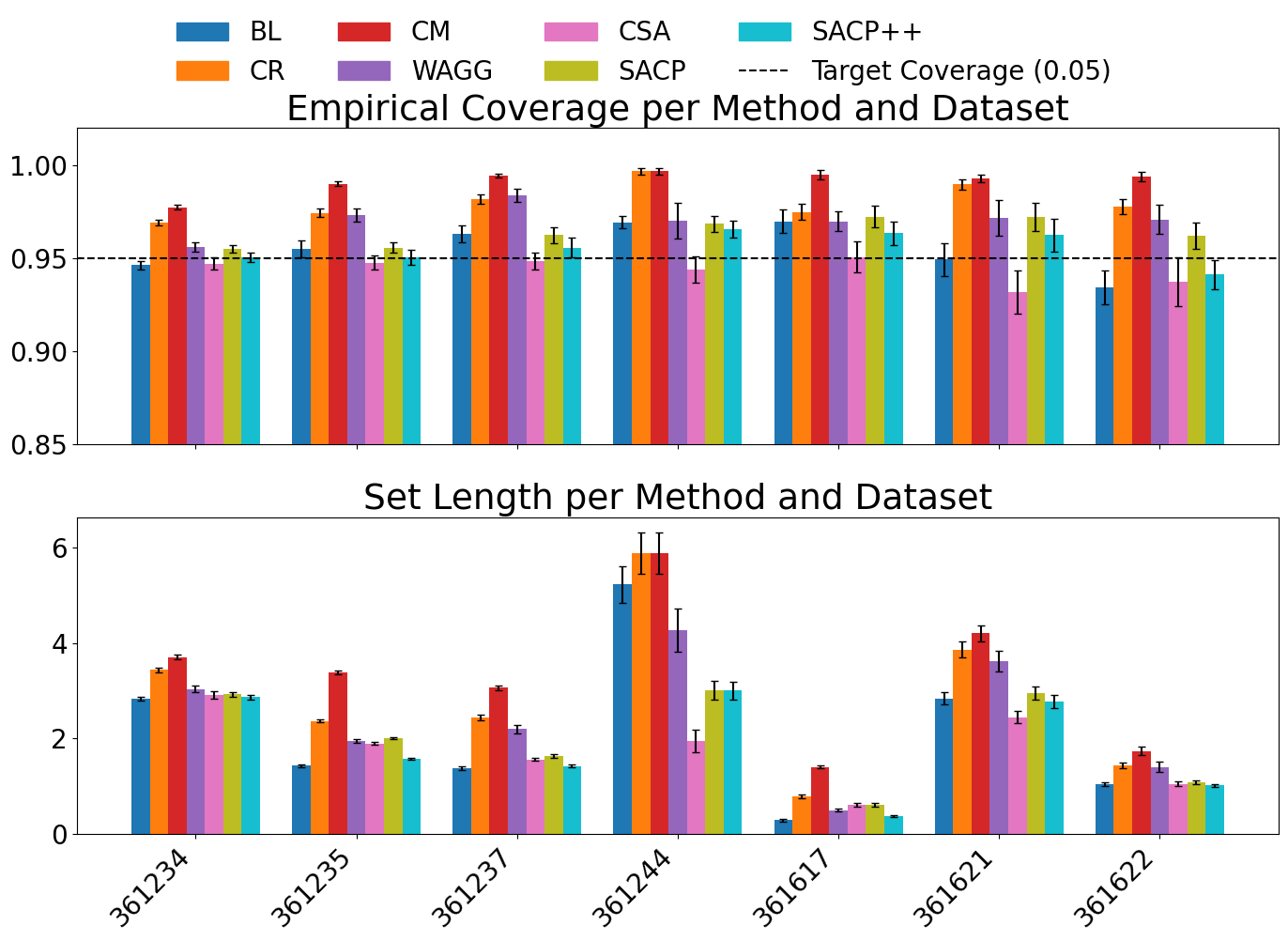}
    \caption{Empirical coverage and average prediction set length per method across OpenML regression datasets.}
    \label{fig:coverage_length_openml20seeds}
\end{figure}
\begin{table*}
  \small
  \centering
  \setlength{\tabcolsep}{1mm} 
  \begin{tabular}{lcccc @{\hspace{7pt}} lcccc}
    \toprule
    \textbf{} &
      \multicolumn{4}{c}{\textbf{CIFAR-10}} &
      \textbf{} &
      \multicolumn{4}{c}{\textbf{MNIST}} \\&
    \textit{Cov }$\alpha{=}0.05$ & \textit{Cov }$\alpha{=}0.1$ &
      \textit{Len }$\alpha{=}0.05$ & \textit{Len }$\alpha{=}0.1$ &&
     \textit{Cov }$\alpha{=}0.05$ & \textit{Cov }$\alpha{=}0.1$ &
      \textit{Len }$\alpha{=}0.05$ & \textit{Len }$\alpha{=}0.1$ \\
    \cmidrule(lr){2-5}\cmidrule(lr){7-10}
    \multicolumn{10}{c}{\textit{Base Models}}\\
    \midrule
    RN56        & 0.949$\pm$0.002 & 0.901$\pm$0.004 & 1.833$\pm$0.062 & 1.352$\pm$0.040 &
    Logistic     & 0.951$\pm$0.004 & 0.902$\pm$0.005 & 1.109$\pm$0.016 & 0.956$\pm$0.007 \\
    ShuffV2     & 0.949$\pm$0.003 & 0.899$\pm$0.004 & 2.322$\pm$0.089 & 1.686$\pm$0.056 &
    RF & 0.951$\pm$0.005 & 0.901$\pm$0.006 & 0.970$\pm$0.005 & 0.908$\pm$0.006 \\
    VGG16       & 0.951$\pm$0.003 & 0.899$\pm$0.004 & 1.531$\pm$0.068 & 1.130$\pm$0.025 &
    HistGB       & 0.950$\pm$0.005 & 0.901$\pm$0.006 & 0.958$\pm$0.005 & 0.903$\pm$0.005 \\
    DLA         & 0.950$\pm$0.003 & 0.899$\pm$0.005 & 1.620$\pm$0.063 & 1.218$\pm$0.037 &
    MLP          & 0.950$\pm$0.003 & 0.908$\pm$0.010 & 0.958$\pm$0.004 & 0.912$\pm$0.010 \\
    EffNet      & 0.949$\pm$0.004 & 0.900$\pm$0.005 & 1.929$\pm$0.067 & 1.366$\pm$0.033 &
    \multicolumn{5}{c}{--} \\
    \midrule
    \multicolumn{10}{c}{\textit{Aggregation Methods}}\\
    \midrule
    BL      & 0.951$\pm$0.003 & 0.899$\pm$0.004 & 1.531$\pm$0.068 & 1.130$\pm$0.025 &
    BL      & 0.950$\pm$0.005 & 0.901$\pm$0.006 & 0.958$\pm$0.005 & 0.903$\pm$0.005 \\
    CR      & 0.948$\pm$0.005 & 0.908$\pm$0.006 & 1.506$\pm$0.054 & 1.194$\pm$0.029 &
    CR      & 0.962$\pm$0.003 & 0.926$\pm$0.005 & 0.975$\pm$0.004 & 0.931$\pm$0.005 \\
    CM      & 0.988$\pm$0.002 & 0.971$\pm$0.003 & 2.089$\pm$0.096 & 1.570$\pm$0.050 &
    CM      & 0.961$\pm$0.018 & 0.967$\pm$0.003 & 0.975$\pm$0.024 & 0.988$\pm$0.004 \\
    Wagg    & 0.948$\pm$0.003 & 0.898$\pm$0.005 & 1.292$\pm$0.028 & 1.032$\pm$0.014 &
    Wagg    & 0.950$\pm$0.004 & 0.901$\pm$0.005 & 0.955$\pm$0.004 & 0.903$\pm$0.005 \\
    CSA     & 0.950$\pm$0.005 & 0.898$\pm$0.007 & 1.294$\pm$0.037 & 1.038$\pm$0.018 &
    CSA     & 0.952$\pm$0.004 & 0.900$\pm$0.006 & 0.958$\pm$0.005 & 0.903$\pm$0.006 \\
    SACP    & 0.950$\pm$0.003 & 0.899$\pm$0.004 & 1.308$\pm$0.033 & 1.034$\pm$0.014 &
    SACP    & 0.951$\pm$0.004 & 0.901$\pm$0.005 & 0.956$\pm$0.004 & 0.904$\pm$0.005 \\
    SACP++  & 0.949$\pm$0.003 & 0.898$\pm$0.004 & \textbf{1.281$\pm$0.027} & \textbf{1.028$\pm$0.011} &
    SACP++  & 0.948$\pm$0.004 & 0.897$\pm$0.005 & \textbf{0.954$\pm$0.004} & \textbf{0.900$\pm$0.006} \\
    \bottomrule
  \end{tabular}
  \caption{Coverage (Cov) and interval length (Len) for CIFAR-10 and MNIST with $\alpha\in\{0.05,0.1\}$: }
  \label{tab:cifar_mnsit}
\end{table*}

\paragraph{Experimental setup.}
For each dataset, we perform 20 random splits into 80\% training, 10\% calibration, and 10\% test sets.  
Inputs and outputs are standardized using their empirical mean and variance.  
In regression, we construct a uniform grid of 255 points between \(\min_{1 \le i \le n} Y_i\) and \(\max_{1 \le i \le n} Y_i\).  

For \textbf{CSA}, we draw \(M = 50\) random projections and perform 20 binary-search iterations to calibrate coverage.  
For \textbf{Wagg}, we conduct a grid search of 200 weight vectors over the \((K-1)\)-simplex.  
For the \textbf{CM} and \textbf{CR} methods, which guarantee coverage of \(1 - 2\alpha\), we set their nominal miscoverage rate to \(\frac{\alpha}{2}\) for a fair comparison.  For
\textbf{SACP++}, we perform a grid search over \(p\) to minimize Equation~\ref{p_opt}.  We use \(K = 7\) diverse base regressors, including linear models, tree-based methods, neural networks, and Bayesian regressors.  
We report the average prediction-set length and empirical coverage on the test set: for regression datasets at \(\alpha = 0.05\), and for classification datasets at both \(\alpha = 0.05\) and \(\alpha = 0.10\).  Further implementation details are provided in Appendix B, and additional experiments, including an analysis of sensitivity to the number of predictors \(K\), are presented in Appendix C.

\paragraph{Results and Discussion.}

The results across all datasets are summarized in Figure~\ref{fig:coverage_length_openml20seeds} and Table~\ref{tab:cifar_mnsit}. In terms of coverage, our proposed methods, \textbf{SACP} and \textbf{SACP++}, consistently achieve the target empirical level across all datasets. Among baselines, \textbf{Wagg} also maintains coverage close to the nominal value, whereas \textbf{CSA} tends to under-cover, and \textbf{CM} and \textbf{CR} frequently exceed the nominal level, resulting in overly conservative (and thus unnecessarily large) prediction sets.  

Regarding prediction-set length, our methods demonstrate strong overall efficiency. 
As expected, \textbf{SACP++} consistently yields shorter sets than \textbf{SACP}. Even the default \textbf{SACP} remains competitive with all baselines, and \textbf{SACP++} is the most efficient method overall. For classification tasks, \textbf{SACP++} reliably produces the smallest prediction sets among all compared methods. In particular, on \textbf{CIFAR-10}, it achieves the lowest variance in prediction-set length across test samples, significantly smaller than that of both the base learners and other aggregation approaches. For regression tasks, \textbf{SACP++} outperforms the best individual model \textbf{BL} on five out of nine datasets and attains the top overall performance among aggregation methods on seven out of nine datasets. These results highlight the benefit of aggregation: by combining predictors through their NCS, our SACP methods maintain valid coverage while producing prediction sets that can be smaller than those of the single predictor with the smallest average set length.

\section{Conclusion}

We proposed SACP, a novel method for aggregating NCSs from \(K\) predictors to construct a single CP set. Our approach transforms calibration and test raw scores into e-variables, which are combined through symmetric functions.  The method applies to any choice of a symmetric aggregation function, while its enhanced variant, SACP++, adaptively selects the function yielding the smallest prediction sets. For strictly monotonic symmetric aggregators, we consider the case when aggregation reduces to a sum of scalar functions, which facilitates both theoretical analysis, via a worst-case bound, and efficient numerical implementation through data-driven aggregation. Empirical results demonstrate that SACP is highly competitive, outperforming state-of-the-art conformal aggregation methods and even the best individual base learner in terms of prediction-set length across all classification datasets and in five out of nine regression datasets.  These findings highlight the effectiveness of conformal aggregation in leveraging shared structure among predictors to improve efficiency while maintaining valid coverage. Future work includes extending SACP++ by exploring symmetric neural architectures capable of learning the optimal aggregation function directly, and investigating how the dependencies among predictors’ uncertainties influence overall performance and efficiency.

\bibliography{aaai2026}

@book{vovk2005algorithmic,
  title={Algorithmic learning in a random world},
  author={Vovk, Vladimir and Gammerman, Alexander and Shafer, Glenn},
  year={2005},
  publisher={Springer}
}

@inproceedings{dietterich2000ensemble,
  title={Ensemble methods in machine learning},
  author={Dietterich, Thomas G},
  booktitle={International workshop on multiple classifier systems},
  pages={1--15},
  year={2000},
  organization={Springer}
}

@article{breiman1996bagging,
  title={Bagging predictors},
  author={Breiman, Leo},
  journal={Machine learning},
  volume={24},
  number={2},
  pages={123--140},
  year={1996},
  publisher={Springer}
}

@misc{angelopoulos2022gentleintroductionconformalprediction,
      title={A Gentle Introduction to Conformal Prediction and Distribution-Free Uncertainty Quantification}, 
      author={Anastasios N. Angelopoulos and Stephen Bates},
      year={2022},
      eprint={2107.07511},
      archivePrefix={arXiv},
      primaryClass={cs.LG},
      url={https://arxiv.org/abs/2107.07511}, 
}

@techreport{krizhevsky2009learning,
  title={Learning multiple layers of features from tiny images},
  author={Krizhevsky, Alex},
  year={2009},
  institution={University of Toronto}
}

@article{lecun2010mnist,
  title={MNIST handwritten digit database},
  author={LeCun, Yann and Cortes, Corinna and Burges, CJ},
  journal={ATT Labs [Online]. Available: http://yann.lecun.com/exdb/mnist},
  volume={2},
  year={2010}
}

@article{vanschoren2014openml,
  title={OpenML: networked science in machine learning},
  author={Vanschoren, Joaquin and van Rijn, Jan N and Bischl, Bernd and Torgo, Luis},
  journal={ACM SIGKDD Explorations Newsletter},
  volume={15},
  number={2},
  pages={49--60},
  year={2014},
  publisher={ACM New York, NY, USA}
}

@misc{gasparin2024merginguncertaintysetsmajority,
      title={Merging uncertainty sets via majority vote}, 
      author={Matteo Gasparin and Aaditya Ramdas},
      year={2024},
      eprint={2401.09379},
      archivePrefix={arXiv},
      primaryClass={stat.ME},
      url={https://arxiv.org/abs/2401.09379}, 
}

@misc{gasparin2024conformalonlinemodelaggregation,
      title={Conformal online model aggregation}, 
      author={Matteo Gasparin and Aaditya Ramdas},
      year={2024},
      eprint={2403.15527},
      archivePrefix={arXiv},
      primaryClass={stat.ML},
      url={https://arxiv.org/abs/2403.15527}, 
}

@misc{yang2024selectionaggregationconformalprediction,
      title={Selection and Aggregation of Conformal Prediction Sets}, 
      author={Yachong Yang and Arun Kumar Kuchibhotla},
      year={2024},
      eprint={2104.13871},
      archivePrefix={arXiv},
      primaryClass={stat.ME},
      url={https://arxiv.org/abs/2104.13871}, 
}

@misc{luo2025weightedaggregationconformityscores,
      title={Weighted Aggregation of Conformity Scores for Classification}, 
      author={Rui Luo and Zhixin Zhou},
      year={2025},
      eprint={2407.10230},
      archivePrefix={arXiv},
      primaryClass={stat.ML},
      url={https://arxiv.org/abs/2407.10230}, 
}

@misc{klein2025multivariateconformalpredictionusing,
      title={Multivariate Conformal Prediction using Optimal Transport}, 
      author={Michal Klein and Louis Bethune and Eugene Ndiaye and Marco Cuturi},
      year={2025},
      eprint={2502.03609},
      archivePrefix={arXiv},
      primaryClass={stat.ML},
      url={https://arxiv.org/abs/2502.03609}, 
}

@misc{braun2025minimumvolumeconformalsets,
      title={Minimum Volume Conformal Sets for Multivariate Regression}, 
      author={Sacha Braun and Liviu Aolaritei and Michael I. Jordan and Francis Bach},
      year={2025},
      eprint={2503.19068},
      archivePrefix={arXiv},
      primaryClass={stat.ML},
      url={https://arxiv.org/abs/2503.19068}, 
}

@misc{rivera2025conformalpredictionensemblesimproving,
      title={Conformal Prediction for Ensembles: Improving Efficiency via Score-Based Aggregation}, 
      author={Eduardo Ochoa Rivera and Yash Patel and Ambuj Tewari},
      year={2025},
      eprint={2405.16246},
      archivePrefix={arXiv},
      primaryClass={stat.ME},
      url={https://arxiv.org/abs/2405.16246}, 
}

@misc{luo2024trustworthyclassificationrankbasedconformal,
      title={Trustworthy Classification through Rank-Based Conformal Prediction Sets}, 
      author={Rui Luo and Zhixin Zhou},
      year={2024},
      eprint={2407.04407},
      archivePrefix={arXiv},
      primaryClass={cs.LG},
      url={https://arxiv.org/abs/2407.04407}, 
}

@article{Abdar_2021,
   title={A review of uncertainty quantification in deep learning: Techniques, applications and challenges},
   volume={76},
   ISSN={1566-2535},
   url={http://dx.doi.org/10.1016/j.inffus.2021.05.008},
   DOI={10.1016/j.inffus.2021.05.008},
   journal={Information Fusion},
   publisher={Elsevier BV},
   author={Abdar, Moloud and Pourpanah, Farhad and Hussain, Sadiq and Rezazadegan, Dana and Liu, Li and Ghavamzadeh, Mohammad and Fieguth, Paul and Cao, Xiaochun and Khosravi, Abbas and Acharya, U. Rajendra and Makarenkov, Vladimir and Nahavandi, Saeid},
   year={2021},
   month=dec, pages={243–297} }

@misc{wang2025aleatoricepistemicexploringuncertainty,
      title={From Aleatoric to Epistemic: Exploring Uncertainty Quantification Techniques in Artificial Intelligence}, 
      author={Tianyang Wang and Yunze Wang and Jun Zhou and Benji Peng and Xinyuan Song and Charles Zhang and Xintian Sun and Qian Niu and Junyu Liu and Silin Chen and Keyu Chen and Ming Li and Pohsun Feng and Ziqian Bi and Ming Liu and Yichao Zhang and Cheng Fei and Caitlyn Heqi Yin and Lawrence KQ Yan},
      year={2025},
      eprint={2501.03282},
      archivePrefix={arXiv},
      primaryClass={cs.AI},
      url={https://arxiv.org/abs/2501.03282}, 
}

@misc{thurin2025optimaltransportbasedconformalprediction,
      title={Optimal Transport-based Conformal Prediction}, 
      author={Gauthier Thurin and Kimia Nadjahi and Claire Boyer},
      year={2025},
      eprint={2501.18991},
      archivePrefix={arXiv},
      primaryClass={stat.ML},
      url={https://arxiv.org/abs/2501.18991}, 
}

@article{vovk2025,
  title={Conformal e-prediction},
  author={Vovk, Vladimir},
  journal={arXiv preprint arXiv:2001.05989},
  year={2025}
}

@article{gauthier2025values,
  title={E-values expand the scope of conformal prediction},
  author={Gauthier, Etienne and Bach, Francis and Jordan, Michael I},
  journal={arXiv preprint arXiv:2503.13050},
  year={2025}
}

@article{gauthier2025backward,
  title={Backward Conformal Prediction},
  author={Gauthier, Etienne and Bach, Francis and Jordan, Michael I},
  journal={arXiv preprint arXiv:2505.13732},
  year={2025}
}

@article{balinsky2024,
  title={Enhancing conformal prediction using e-test statistics},
  author={Balinsky, Alexander A and Balinsky, Alexander D},
  journal={arXiv preprint arXiv:2403.19082},
  year={2024}
}

@article{ramdas2024hypothesis,
  title={Hypothesis testing with e-values},
  author={Ramdas, Aaditya and Wang, Ruodu},
  journal={arXiv preprint arXiv:2410.23614},
  year={2024}
}

@article{Merg_e_val,
  title={E-values: Calibration, combination and applications},
  author={Vovk, Vladimir and Wang, Ruodu},
  journal={The Annals of Statistics},
  volume={49},
  number={3},
  pages={1736--1754},
  year={2021},
  publisher={Institute of Mathematical Statistics}
}

@article{wang2024only,
  title={The only admissible way of merging e-values},
  author={Wang, Ruodu},
  journal={arXiv preprint arXiv:2409.19888},
  year={2024}
}

@article{wang2022false,
  title={False discovery rate control with e-values},
  author={Wang, Ruodu and Ramdas, Aaditya},
  journal={Journal of the Royal Statistical Society Series B: Statistical Methodology},
  volume={84},
  number={3},
  pages={822--852},
  year={2022},
  publisher={Oxford University Press}
}

@article{gasparin2025combining,
  title={Combining exchangeable p-values},
  author={Gasparin, Matteo and Wang, Ruodu and Ramdas, Aaditya},
  journal={Proceedings of the National Academy of Sciences},
  volume={122},
  number={11},
  pages={e2410849122},
  year={2025},
  publisher={National Academy of Sciences}
}

@misc{zaheer2018deepsets,
      title={Deep Sets}, 
      author={Manzil Zaheer and Satwik Kottur and Siamak Ravanbakhsh and Barnabas Poczos and Ruslan Salakhutdinov and Alexander Smola},
      year={2018},
      eprint={1703.06114},
      archivePrefix={arXiv},
      primaryClass={cs.LG},
      url={https://arxiv.org/abs/1703.06114}, 
}

@book{vincent1912functions,
  title={The Functions of the Vibrissae in the Behavior of the White Rat...},
  author={Vincent, Stella Burnham},
  volume={1},
  year={1912},
  publisher={University of Chicago}
}

@misc{jin2023selectionpredictionconformalpvalues,
      title={Selection by Prediction with Conformal p-values}, 
      author={Ying Jin and Emmanuel J. Candès},
      year={2023},
      eprint={2210.01408},
      archivePrefix={arXiv},
      primaryClass={stat.ME},
      url={https://arxiv.org/abs/2210.01408}, 
}

@article{thomas1980appropriate,
  title={On appropriate procedures for combining probability distributions within the same family},
  author={Thomas, Ewart AC and Ross, Brian H},
  journal={Journal of Mathematical Psychology},
  volume={21},
  number={2},
  pages={136--152},
  year={1980},
  publisher={Elsevier}
}

@article{blanchet2024convolution,
  title={Convolution bounds on quantile aggregation},
  author={Blanchet, Jose and Lam, Henry and Liu, Yang and Wang, Ruodu},
  journal={Operations Research},
  year={2024},
  publisher={INFORMS}
}

@article{lichtendahl2013better,
  title={Is it better to average probabilities or quantiles?},
  author={Lichtendahl Jr, Kenneth C and Grushka-Cockayne, Yael and Winkler, Robert L},
  journal={Management Science},
  volume={59},
  number={7},
  pages={1594--1611},
  year={2013},
  publisher={INFORMS}
}

@article{vovk2022admissible,
  title={Admissible ways of merging p-values under arbitrary dependence},
  author={Vovk, Vladimir and Wang, Bin and Wang, Ruodu},
  journal={The Annals of Statistics},
  volume={50},
  number={1},
  pages={351--375},
  year={2022},
  publisher={Institute of Mathematical Statistics}
}

@article{vovk2020combining,
  title={Combining p-values via averaging},
  author={Vovk, Vladimir and Wang, Ruodu},
  journal={Biometrika},
  volume={107},
  number={4},
  pages={791--808},
  year={2020},
  publisher={Oxford University Press}
}

@article{gasparin2025improving,
  title={Improving the statistical efficiency of cross-conformal prediction},
  author={Gasparin, Matteo and Ramdas, Aaditya},
  journal={arXiv preprint arXiv:2503.01495},
  year={2025}
}

@book{dhaene2002a,
  title={The Concept of Comonotonicity in Actuarial Science and Finance: Theory},
  author={Dhaene, Jan and Denuit, Michel and Goovaerts, Marc and Kaas, Rob and Vyncke, Dries},
  year={2002},
  publisher={Springer Science \& Business Media}
}

@article{dhaene2002b,
  title={{The Concept of Comonotonicity in Actuarial Science and Finance: Applications}},
  author={Dhaene, Jan and Denuit, Michel and Goovaerts, Marc J and Kaas, Rob and Vyncke, Dries},
  journal={Insurance: Mathematics \& Economics},
  volume={31},
  number={2},
  pages={133--161},
  year={2002},
  publisher={Elsevier}
}

@inproceedings{Dheur2025-unified,
  title = {A unified comparative study with generalized conformity scores for multi-output conformal regression},
  author = {Dheur, Victor and Fontana, Matteo and Estievenart, Yorick and Desobry, Naomi and Ben Taieb, Souhaib},
  booktitle = {The 42nd International Conference on Machine Learning},
  year = {2025},
}

\appendix
%\newpage
\onecolumn
\clearpage
\large
\section{Proofs of results}
\label{appendix:proofs}
\paragraph{Notation.}
For any vector $X=(X_1,\dots,X_n)$, write $X_{(1)}\le\cdots\le X_{(n)}$ for order statistics; 
$|A|$ is the cardinality of a set $A$; $\lfloor\cdot\rfloor$ and $\lceil\cdot\rceil$ denote floor and ceiling.

All the predictors $\hat \mu^{(1)},\cdots,\hat \mu^{(K)}$ are trained on $\{(X_i,Y_i)\}_{i\in \mathcal{I}_{train}}$ and are thus considered to be deterministic functions, so the calibration scores are functions of the $(X_i,Y_i)$, for all $i\in \mathcal{I}_{cal} = \{1,\dots,n\}.$ Denote $Z =( Z_1,\dots,Z_n,Z_{n+1}) $ with $Z_i:=(X_i,Y_i), Z_{n+1} := (X_{test},Y_{test})$, and $Z_\pi := (Z_{\pi(1)},\dots,Z_{\pi(n+1)}) $ for any permutation $\pi.$ Denote $\mathcal S_K$ the groups of permutations of $\{1,\dots,K\}$.

\paragraph{Model-permutation invariance.}
Consider a function $f:\mathbb{R}^K \to \mathbb{R}$ used to aggregate the
per-model scores. The labels
$1,\dots,K$ of the base predictors carry no intrinsic meaning; relabeling the
models should not change the aggregated score. Formally, for any permutation
$\sigma \in \mathcal S_K$, we require
\begin{equation}
    f\bigl(x_{(1)},\dots,x_{(K)}\bigr)
    \;=\;
    f\bigl(x_{(\sigma(1))},\dots,x_{(\sigma(K))}\bigr).
\end{equation}
This exactly means that $f$ is
a symmetric function of its arguments. This restriction is thus relevant in our context, and also facilitates the theoretical study and numerical implementation of the SACP method as explained in the paper.

\paragraph{Proof of Theorem \ref{main_theorem}}

\begin{proof}

For any vector $z=(z_1,\dots,z_{n+1})$ and any $i\in\{1,\dots,n+1\}$, define the
$K$-dimensional vector 
\[
\mathbf{E}_i(z) := \left(
\frac{s^{(k)}(z_i)}{\frac{1}{n+1}\sum_{j=1}^{n+1} s^{(k)}(z_j)}
\right)_{1\le k\le K}
\in\mathbb{R}^K,
\]
and let
\[
F_i(z) := f(\mathbf{E}_i(z)), \qquad i=1,\dots,n+1,
\]
where $f:\mathbb{R}^K\to\mathbb{R}$ is a fixed measurable  function. The aggregated scores used by SACP are then $F_i := F_i(Z)$.

Let $\pi$ be any permutation of $\{1,\dots,n+1\}$, and denote
$Z_\pi := (Z_{\pi(1)},\dots,Z_{\pi(n+1)})$.
Since the denominator in the definition of $\mathbf{E}_i(z)$ is a sum over all indices
$j$, it is invariant under permutation. Moreover, the $i$-th element of $Z_\pi$ is $Z_{\pi(i)}$. Hence, for every $i$,
\[
\mathbf{E}_i(Z_\pi)
=
\left(
\frac{s^{(k)}(Z_{\pi(i)})}{\frac{1}{n+1}\sum_{j=1}^{n+1} s^{(k)}(Z_{\pi(j)})}
\right)_{k=1}^K
=
\left(
\frac{s^{(k)}(Z_{\pi(i)})}{\frac{1}{n+1}\sum_{j=1}^{n+1} s^{(k)}(Z_j)}
\right)_{k=1}^K
=
\mathbf{E}_{\pi(i)}(Z),
\]
and therefore
\[
F_i(Z_\pi) = f(\mathbf{E}_i(Z_\pi)) = f(\mathbf{E}_{\pi(i)}(Z)) = F_{\pi(i)}(Z).
\]
Thus the map $Z \mapsto (F_1(Z),\dots,F_{n+1}(Z))$ is permutation equivariant in the
sample index. Since $Z\overset{d}{=}Z_\pi$ for all $\pi$, it follows that
$(F_1,\dots,F_{n+1})$ is an exchangeable sequence. Therefore, we are exactly in the standard
conformal prediction setting, with nonconformity scores given by
$\{F_i(y)\}_{i=1}^{n+1}$.

\end{proof}

\paragraph{Proof of Proposition \ref{prop:V_ordering}}

\begin{proof}
Let \(x_1,\dots,x_n\) be real numbers and write their order statistics as
$x_{(1)}\le x_{(2)}\le\cdots\le x_{(n)}$, and 
let \(g\colon\mathbb{R}\to\mathbb{R}\) be any strictly decreasing function, and define

$$\tilde x_i \;=\; g(x_i),
\qquad
\tilde x_{(1)}\le \tilde x_{(2)}\le\cdots\le \tilde x_{(n)}$$
their increasing order statistics. If $g$ is increasing, it preserves the order, so the $\tilde{x}_{(k)} = g(x_{(k)})$. Otherwise, \(g\) reverses the order,
\begin{equation}\label{eq:reverse_order}
g(x_{(k)}) \;=\; \tilde x_{(n-k+1)},
\qquad k=1,\dots,n.
\end{equation}
In particular, $g(x_{(\,\lceil (1-\alpha)(n+1)\rceil)} 
)\;=\;
\tilde x_{(\,\lfloor \alpha(n+1)\rfloor)}$
because 
\begin{align}
  m\;=\;\bigl\lceil(1-\alpha)(n+1)\bigr\rceil\Longrightarrow
m-1 < (1-\alpha)(n+1)\le m
\;\Longrightarrow\;
n+1-m \le \alpha(n+1)< n+2-m
\end{align}
hence

$$\bigl\lfloor\alpha(n+1)\bigr\rfloor = n+1-m
= n - \bigl\lceil(1-\alpha)(n+1)\bigr\rceil + 1.$$
\end{proof}

\paragraph{Proof of Proposition \ref{prop:rho_composition}}
\begin{proof}
    
Consider 
\begin{equation}
    f(\mathbf{x}) = \rho\circ\Phi(\mathbf{x})
\end{equation}
with $\rho,\phi$ strictly monotonic 
as mentioned in the paper. The aggregated calibration scores are given by $\{F_i\}_{1\le i\le n}=\{\rho\circ\Phi (\mathbf{E}_i)\}_{1\le i\le n}$, following the same notations as in the paper: $\mathbf{E}_i = (E_i^{(1)},\dots,E_i^{(K)}), i=1,\dots,n.$

Denote $V_i:=\Phi(\mathbf{E}_i)= \sum_{k=1}^K\phi(E^k_i)$ the merged scores with $\Phi$, and $k=\lceil(1-\alpha)(n+1)\rceil.$ It follows from the previous proposition:

\begin{itemize}
    \item If $\rho$ is strictly increasing 
$$ \mathcal{C}_\alpha^{\rho\circ \Phi,\,\uparrow}(X_{test})
=\{y \mid \rho\circ\Phi(E_{test}(y))\leq \rho( V_{(k)} )\}  $$
$$=\{y \mid \Phi(E_{test}(y))\leq V_{(k)}) \} =  \mathcal{C}_\alpha^{ \Phi,\,\uparrow}(X_{test})$$

\item If $\rho$ is strictly decreasing  
$$ \mathcal{C}_\alpha^{\rho\circ \Phi,\,\uparrow}(X_{test})
=\{y \mid \rho\circ\Phi(E_{test}(y))\leq \rho(V_{(n-k+1)}) \}  $$
$$=\{y \mid \Phi(E_{test}(y))\geq V_{(n-k+1)}) \} =  \mathcal{C}_\alpha^{ \Phi,\,\downarrow}(X_{test}).$$
\end{itemize}

\end{proof}

We are now ready to prove the main theorem of the paper. But before that we should establish a useful bound on the empirical quantile of the aggregated score $\hat Q_\alpha$, starting with a lemma.

\begin{lemma}[Empirical quantile upper bound]
Fix $K>1$ and $n\in\mathbb N$. Let $\alpha\in[\frac{K}{n+1},1)$. For  $k=1,\dots,K$ let
$(S^{(k)}_{1},\dots,S^{(k)}_{n})\in\mathbb R^n$ be real numbers. Consider the 'aggregated vector' $(V_1,\dots,V_n)$ where $$V_i:=\sum_{k=1}^KS^{(k)}_i$$ 
and define the empirical quantiles : 
$$ \hat Q_{\alpha/K}^k:=S^{(k)}_{(\lceil(n+1)(1-\alpha/K)\rceil)},\quad \hat Q_\alpha := V_{(\lceil(n+1)(1-\alpha)\rceil)}.$$
Then \begin{equation}
    \hat Q_{\alpha} \le \sum_{k=1}^K\hat Q_{\alpha/K}^k
\end{equation}
\end{lemma}

\begin{proof}
Set $E_k:=\{\,i:\ S^{(k)}_{i}\le \hat Q_{\alpha/K}^k\,\}$ and $F:=\{\,i:\ V_{i}\le  \sum_{k=1}^K \hat Q^k_{\alpha/K}\,\}$.
By definition of the quantile, $|E_k|\ge\lceil(n+1)(1-\alpha/K)\rceil$. 
Hence, by the union bound on sets,
\begin{align}
    \big|\bigcap_{k=1}^K E_k\big|
=n-|\bigcup_{k=1}^K E_k^c| \ge\ n-\sum_{k=1}^K |E_k^c|\\
\ge n-\sum_{k=1}^K (n-\lceil(n+1)(1-\alpha/K)\rceil)\\
\ge n+ \sum_{k=1}^K((n+1)(1-\alpha/K)-n)\\
\ge n+K-(n+1)\alpha\ge (n+1)(1-\alpha).
\end{align}

For every $i\in\bigcap_{k=1}^K E_k$ we have 
$V_i\le \sum_{k=1}^K \hat Q_{\alpha/K}^k$. Thus

\begin{align}
    \big|F|\ge \big |\bigcap_{k=1}^K E_k\big|\ge(n+1)(1-\alpha).
\end{align}
Therefore, since the cardinal of set is an integer we get $|F|\ge \lceil(n+1)(1-\alpha)\rceil.$ By definition of the quantile, we conclude that
$$\hat Q_{\alpha} \le \sum_{k=1}^K\hat Q_{\alpha/K}^k.$$

\end{proof}

The lemma above can be used to upper bound the empirical quantile \(\widehat Q_\alpha(y)\) of the SACP method. 
Let \(\Phi\in\mathcal{F}_{\text{agg}}\), and for each \(k=1,\dots,K\) denote by \(\widehat Q^k_\beta\) the empirical \((1-\beta)\)-quantile of the raw nonconformity scores \(s^{(k)}(X_i,Y_i)\), \(i=1,\dots,n\).
For a fixed candidate label \(y\), note that the e-value normalization
\[
x \;\mapsto\; \frac{(n+1)\,x}{\sum_{i=1}^n s^{(k)}(X_i,Y_i)\;+\;s^{(k)}(X_{\text{test}},y)}
\]
has a denominator that does not depend on \(i\); hence it preserves the per-\(k\) ordering of the scores. If, in addition, \(\phi\) is nondecreasing, applying \(\phi\) also preserves this ordering. Therefore the transformed per-model scores are
\[
\tilde s_i^{(k)}(y)
\;:=\;
\phi\!\left(
\frac{(n+1)\,s^{(k)}(X_i,Y_i)}
{\sum_{j=1}^n s^{(k)}(X_j,Y_j)\;+\;s^{(k)}(X_{\text{test}},y)}
\right),
\quad i=1,\dots,n,
\]
and the aggregated scores are \(F_i(y)=\sum_{k=1}^K \tilde s_i^{(k)}(y)\).
Applying the previous lemma to the \(K\) vectors \(\{\tilde s_i^{(k)}(y)\}_{i=1}^n\) yields
\begin{equation}
\widehat{Q}_\alpha(y)
\;=\;
F_{(\lceil(1-\alpha)(n+1)\rceil)}(y)
\;\le\;
\sum_{k=1}^K \tilde s^{(k)}_{(\lceil(1-\alpha/K)(n+1)\rceil)}(y)
\;=\;
\sum_{k=1}^K
\phi\!\left(
\frac{(n+1)\,\widehat Q^k_{\alpha/K}}
{\sum_{i=1}^n s^{(k)}(X_i,Y_i)\;+\;s^{(k)}(X_{\text{test}},y)}
\right).
\label{quantile_bound_alpha/K}
\end{equation}

\noindent\emph{Reader’s note.}
Stronger bounds are possible under additional structure (e.g., when the per-model score vectors are comonotonic \cite{dhaene2002a,dhaene2002b}), but such assumptions are typically too strong to be imposed in practice.

\subsection{Proof of Theorem \ref{theorem : bound C^f_alpha}}
We have now eliminated the dependence on \(\rho\) and obtained an explicit upper bound on the empirical quantile \(\widehat Q_\alpha(y)\) of the aggregated scores. This reduction allows us to study the SACP prediction set. We adopt the same notations as in  Theorem \ref{main_theorem} and Section \ref{sec:SACP}. Consider an aggregation function $\Phi \in \mathcal{F}_{\text{agg}}$ that is strictly non-decreasing in each coordinate. Our goal is to control the size of $$C_\alpha^{\Phi,\uparrow}:=\{y\in\mathcal{Y}\mid F_{test}(y)\leq \hat{Q}_\alpha(y)\}.$$

\begin{figure}[H]
    \centering

\begin{tikzpicture}[
  node distance=7mm and 10mm,
  box/.style={rectangle, draw, rounded corners=2pt, align=center,
              inner sep=3pt, minimum width=32mm, minimum height=8mm, font=\small},
  arrow/.style={-{Latex}, thick}
]
\node[box] (s3a) {study of \\$g(y):=F_{\text{test}}(y)-\widehat Q_\alpha(y)$};
\node[box, right=of s3a] (s3) {Construct \\
$U(y)\ge\hat Q_\alpha(y),\quad L(y)\le F_{test}(y)$};
\node[box, right=of s3] (s4) {bound $y^*$ s.t.\\
$U(y^*)=L(y^*)$};
\node[box, right=of s4] (s6) {Final bound on $|\mathcal C_\alpha^\Phi|$};

\draw[arrow] (s3a) -- (s3);
\draw[arrow] (s3) -- (s4);

\draw[arrow] (s4) -- (s6);
\end{tikzpicture}
\caption{Proof roadmap for bounding SACP prediction set}
    \label{schema_preuve}
\end{figure}
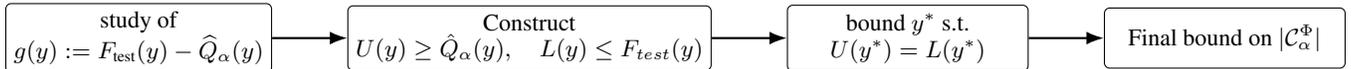

Recall that we seek a \emph{(worst–case)} upper bound on the width of the aggregated prediction set that holds for all admissible choices of $\phi$.
This bound characterizes the maximal width any aggregated set can attain; hence any
particular choice of $\phi$ can only tighten the set relative to this envelope.
In what follows we restrict attention to strictly increasing $\phi$; the strictly decreasing case
is analogous (the inequalities reverse and yields the same bound).
The main steps of the proof are displayed in  Fig.~\ref{schema_preuve}.

We start by studying the nature of the functions that appear in the prediction set. The following lemmas will help us achieve that :
\begin{lemma}
    Let $\{s_1(y),...,s_n(y)\}$ be continuous functions on $[0,+\infty)$. Then, for $k\in\{1,..,n\}$, their $k$-th order statistic is also a continuous function of $y$.
\end{lemma}

\begin{proof}
    The $k$-th order statistic $s_{(k)}(y)$ can be written as
$s_{(k)}(y) = \max_{\substack{J \subseteq \{1, \ldots, n\} \\ |J| = n-k+1}} \left( \min_{j \in J} s_j(y) \right)$ and the max and min functions are continuous.
\end{proof}

\begin{lemma}
\label{lemma:quantile_incre}
    If there exists a $y_0$ such that the functions $\{s_1(y),...,s_n(y)\}$ are non-decreasing in $[y_0,+\infty)$ , then their $k$-th quantile is also non decreasing on $[y_0,+\infty)$.
\end{lemma}

\begin{proof}
   Fix \(y_1,y_2\in[y_0,\infty)\) with \(y_1<y_2\).  Then for each \(i\),
\[
s_i(y_1)\;\le\;s_i(y_2).
\]
Let 
$s_{(1)}(y_1)\le s_{(2)}(y_1)\le\cdots\le s_{(n)}(y_1)$ and 
$s_{(1)}(y_2)\le s_{(2)}(y_2)\le\cdots\le s_{(n)}(y_2)$
be the order statistics at \(y_1\) and \(y_2\), respectively.  Since for each \(i\), we have \(s_i(y_1) \le s_i(y_2)\), it follows that
\[
s_{(k)}(y_1)\;\le\;s_{(k)}(y_2), \qquad k=1,\dots,n.
\]
\end{proof}

\begin{lemma}
    
For a test sample $X_{test}$, denote 
\begin{equation}
    y_0 = \min_{1\leq k\leq K} \hat{\mu}^{(k)}(X_{test}), \quad  
y_1 = \max_{1\leq k\leq K} \hat{\mu}^{(k)}(X_{test}). 
\end{equation}
Then $y\to\hat{Q}_\alpha(y)$ is  continuous strictly increasing on $(-\infty,y_0]$, and strictly decreasing on  $[y_1,+\infty)$. On the contrary, $F_{test}(y)$ is strictly decreasing  on  $(-\infty,y_0]$ , and strictly increasing on $[y_1,+\infty)$. 
In particular, both of these functions are bijective on these two intervals.
\end{lemma}

%Additionally, the Lipshitz constant of $y\to E^{(k)}_{test}(y)$ and $y\to E^{(k)}_i$ are $c_k=\frac{n+1}{\sum_{i=1}^n s^{(k)}_i}$ and $b_{k,i}=\frac{(n+1)s^{(k)}_j}{(\sum_{j=1}^n s^{(k)}_j)^2}\leq c_k$, respectively.

\begin{proof}
The continuous function $$y\to \frac{s^{(k)}(X_i,Y_i)}{(\sum_{i=1}^ns^{(k)}(X_i,Y_i) + |\hat{\mu}^{(k)}(X_{test})-y|)/n+1}$$ is strictly increasing on $(-\infty,\hat{\mu}^{(k)}(X_{test})]$, so in particular $(-\infty,y_0]$. It is also strictly decreasing on $[y_1,+\infty)$. Because $\Phi$ is increasing, the aggregated score $F_i:=\Phi(E^1_i,\dots,E_i^{(K)})$ behaves the same, and we get the result using lemma \ref{lemma:quantile_incre}. By the same logic, the continuous function $$y\to \frac{|\hat{\mu}^{(k)}(X_{test})-y|}{(\sum_{i=1}^ns^{(k)}(X_i,Y_i) + |\hat{\mu}^{(k)}(X_{test})-y|)/n+1}$$ is strictly increasing on $[y_1,+\infty)$ and strictly decreasing on $(-\infty,y_0]$, which gives the behavior of $F_{test}.$

\end{proof}

\paragraph{Endpoints study.}
If $\mathcal C_\alpha^\Phi(X_{\mathrm{test}})=\varnothing$, the length bound is trivial.
If $C_\alpha^\Phi$ is unbounded, it is not useful in practice; in what follows we work on the (typical) event that $C_\alpha^\Phi$ is nonempty and bounded.
Define
\[
g(y):=F_{\text{test}}(y)-\hat Q_\alpha(y),\qquad
\mathcal C_\alpha^\Phi(X_{\text{test}}):=\{y\in\mathbb R:\ g(y)\le 0\}=g^{-1}((-\infty,0]).
\]
By the previous lemmas, $g$ is continuous and strictly decreasing on $(-\infty,y_0]$ and
strictly increasing on $[y_1,\infty)$.
Since $g$ is continuous and $(-\infty,0]$ is closed, $C_\alpha^\Phi$ is closed (hence a union
of closed intervals). Consider
\[
y_{\text{left}}:=\inf \mathcal C_\alpha^f(X_{\text{test}}),\qquad
y_{\text{right}}:=\sup \mathcal C_\alpha^f(X_{\text{test}}),
\]
such that we have $$\mathcal C_\alpha^\Phi(X_{\text{test}})\subset [\,y_{\text{left}},\,y_{\text{right}}\,],\qquad
g(y_{\text{left}})=0\quad\text{and}\quad g(y_{\text{right}})=0
$$
by continuity of $g$.

The conformal prediction set of our base predictors are all centered in $\hat{\mu}^{(k)}(X_{test})$ and of length equal to the empirical quantile. Intuitively, we can expect at least some of the centers $\hat{\mu}^{(k)}(X_{test})$ to belong to $\mathcal{C}_\alpha^\Phi(X_{\text{test}})$. This is what is shown in the following proposition: 
\begin{proposition}
Either $y_0\in \mathcal{C}_\alpha^\Phi(X_{\text{test}})$ or $y_0\le y_{\mathrm{left}}$. Likewise, either
$y_1\in \mathcal{C}_\alpha^\Phi(X_{\text{test}})$ or $y_1\ge y_{\mathrm{right}}$.
\end{proposition}

\begin{proof}
Recall $g(y)=F_{\mathrm{test}}(y)-\widehat Q_\alpha(y)$ and
$\mathcal{C}_\alpha^\Phi(X_{\text{test}})=\{y:\,g(y)\le 0\}$, with $g$ continuous, strictly decreasing on
$(-\infty,y_0]$ and strictly increasing on $[y_1,\infty)$, and
$g(y_{\mathrm{left}})=g(y_{\mathrm{right}})=0$.

Suppose $y_0\notin \mathcal{C}_\alpha^\Phi$ and $y_0>y_{\mathrm{left}}$.
Then $g(y_0)>0$ while $g(y_{\mathrm{left}})=0$ and $y_{\mathrm{left}}<y_0$.
Since $g$ is strictly decreasing on $(-\infty,y_0]$, we have
$g(y_{\mathrm{left}})>g(y_0)>0$, a contradiction to $g(y_{\mathrm{left}})=0$.
Hence, if $y_0\notin \mathcal{C}_\alpha^\Phi(X_{\text{test}})$, necessarily $y_0\le y_{\mathrm{left}}$. A similar argument proves the statement of the proposition for $y_1$.
\end{proof}

\begin{figure}[H]
  
\begin{center}
\begin{tikzpicture}[scale=1.5]
  \draw[->] (0,0) -- (5,0) node[right] {};
   \node[anchor=north west, scale=0.8] at (0,0.4) {$A$ =\{ $y_0 \le y_{\text{left}} \le y_{\text{right}} \le y_1$\}};
  \draw[fill=black] (1,0) circle (0.04) node[below] {$y_0$};
  \draw[fill=blue] (2,0) circle (0.04) node[below] {\textcolor{blue}{$y_{\text{left}}$}};
  \draw[fill=blue] (4,0) circle (0.04) node[below] {{\textcolor{blue}{$y_{\text{right}}$}}};
  \draw[fill=black] (4.5,0) circle (0.04) node[below] {$y_{1}$};

  \draw[thick, blue] (2,0.0) -- (4,0.0);
  \node[above, blue] at (3,0.15) {};
\end{tikzpicture}
\end{center}

\begin{center}
\begin{tikzpicture}[scale=1.5]
  \draw[->] (0,0) -- (5,0) node[right] {};
  \node[anchor=north west, scale=0.8] at (0,0.4) {$B_1$ =\{ $y_0<y_{\text{left}} < y_1 < y_{\text{right}}$\}};
  \draw[fill=black] (1,0) circle (0.04) node[below] {$y_0$};
  \draw[fill=blue] (2,0) circle (0.04) node[below] {\textcolor{blue}{$y_{\text{left}}$}};
  \draw[fill=blue] (4,0) circle (0.04) node[below] {{\textcolor{blue}{$y_{\text{right}}$}}};
  \draw[fill=black] (3.5,0) circle (0.04) node[below] {$y_{1}$};
  
  \draw[thick, blue] (2,0.0) -- (4,0.0);
  \node[above, blue] at (3,0.15) {};
\end{tikzpicture}
\end{center}

\begin{center}
\begin{tikzpicture}[scale=1.5]
  \draw[->] (0,0) -- (5,0) node[right] {};
  \node[anchor=north west, scale=0.8] at (0,0.4) {$B_2$ = \{ $y_{\text{left}} < y_0 <y_{\text{right}}<y_1$\}};
  \draw[fill=black] (2.5,0) circle (0.04) node[below] {$y_0$};
  \draw[fill=blue] (2,0) circle (0.04) node[below] {\textcolor{blue}{$y_{\text{left}}$}};
  \draw[fill=blue] (4,0) circle (0.04) node[below] {{\textcolor{blue}{$y_{\text{right}}$}}};
  \draw[fill=black] (4.5,0) circle (0.04) node[below] {$y_{1}$};
  
  \draw[thick, blue] (2,0.0) -- (4,0.0);
  \node[above, blue] at (3,0.15) {};
\end{tikzpicture}
\end{center}
\begin{center}
\begin{tikzpicture}[scale=1.5]
  \draw[->] (0,0) -- (5,0) node[right] {};
  \node[anchor=north west, scale=0.8] at (0,0.4) {$C$ =\{ $y_{\text{left}} < y_0<y_1 <y_{\text{right}}$\}};
  \draw[fill=black] (2.5,0) circle (0.04) node[below] {$y_0$};
  \draw[fill=blue] (2,0) circle (0.04) node[below] {\textcolor{blue}{$y_{\text{left}}$}};
  \draw[fill=blue] (4,0) circle (0.04) node[below] {{\textcolor{blue}{$y_{\text{right}}$}}};
  \draw[fill=black] (3.5,0) circle (0.04) node[below] {$y_{1}$};
  
  \draw[thick, blue] (2,0.0) -- (4,0.0);
  \node[above, blue] at (3,0.15) {};
\end{tikzpicture}
\end{center}
\caption{Cases of where $y_0,y_1$ lie in or out of $C^f_\alpha$. }
    \label{Cases}
\end{figure}
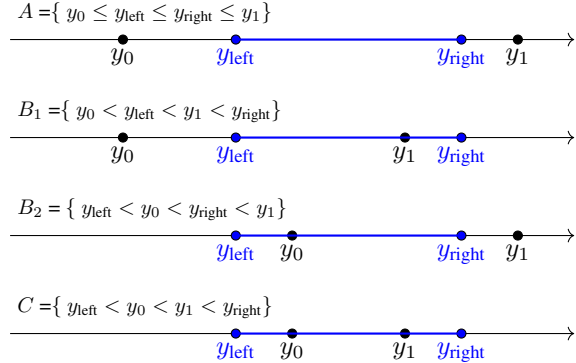

We shall now distinguish between all possible cases, presented in Figure \ref{Cases}. 
\vspace{-10pt}
\noindent We denote the events $A,B_1,B_2,C$ of the respective cases in the figure. In the event $A$ we can simply bound the length of our prediction set  with $y_1 - y_0 = \Delta_{\text{test}}$. 
For the events $A,B_1,B_2$ we establish a common bound. This is achieved with the following result. 

\begin{proposition}
   If $y_0\in \mathcal{C}_\alpha^\Phi(X_{\text{test}})$, then $y_{left}\geq y_0-\max_k\hat{Q}_{\alpha/K}^k.$ If $y_1\in \mathcal C^\Phi_\alpha(X_{\text{test}})$, then
$y_{right}\leq y_1+\max_k\hat{Q}_{\alpha/K}^k$.
\end{proposition}
\begin{proof}
    The proof relies on upper bounding the quantile function $\hat{Q}_\alpha^k(y)$ by $U(y)$  and  lower bounding $F_{test}(y)$ by $L(y)$ on $(-\infty,y_0]$. If $U,L$ are also bijective, they also uniquely intersect at some point $y^*$. Thus, necessarily $y^*\leq y_{left}.$

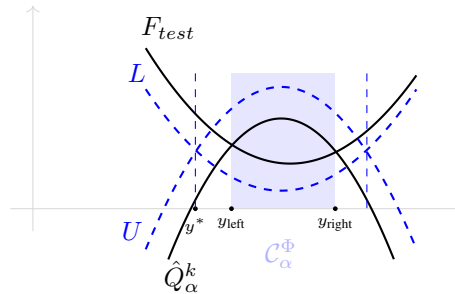
\begin{figure}[H]
    \centering
        \begin{center}
    \begin{tikzpicture}[scale=0.6]
\fill[blue!10] (0.9,0) rectangle (3.2,3 ) node at (2,-1) {\textcolor{blue!30}{$\mathcal{C}_\alpha^\Phi$}};
  % Light axes
  \draw[->, gray!30] (-4,0) -- (6,0);
  \draw[->, gray!30] (-3.5,-0.5) -- (-3.5,4.5);

  % Main concave (T_alpha)
  \draw[domain=-0.5:4.5, smooth, samples=200,  thick, black]
    plot (\x, {-(0.5)*(\x-2)^2 + 2});
  \node at (-0.2,-1.5) {$\hat{Q}^k_\alpha$};

  % Main convex (tilde E)
  \draw[domain=-1:5, smooth, samples=200, thick, black]
    plot (\x, {(0.5)*(\x-2.2))^2 + 1});
  \node at (-0.5,4) {$F_{test}$};

  % Upper concave (U), blue dashed, wider and above T_alpha
  \draw[domain=-1:5, smooth, samples=200, thick, blue, dashed]
    plot (\x, {-(0.4)*(\x-2)^2 + 2.7});
  \node[blue] at (-1.3,-0.5) {$U$};

  % Lower convex (L), blue dashed, wider and below tilde E
  \draw[domain=-1:5, smooth, samples=200, thick, blue, dashed]
    plot (\x, {(0.25)*(\x-2)^2 + 0.4});
  \node[blue] at (-1.2,3) {$L$};
\fill[black] (0.9,0) circle (1.5pt) node[below] {\tiny$ y_{\text{left}}$};
 
 \fill[black] (0.1,0) circle (1.5pt) node[below] {\tiny$ y^*$};

  \fill[black] (3.2,0) circle (1.5pt) node[below] {\tiny$y_{\text{right}}$};
\draw[-,blue, dashed] (0.1,3) -- (0.1,0);

 \draw[-,blue, dashed] (3.9,3) -- (3.9,0);

\end{tikzpicture}
\end{center}
    \caption{An illustration of $U,L$ maps intersection points.}
    \label{fig:placeholder}
\end{figure}
\noindent Let us determine $U(y)$ and $L(y)$. By definition, for $y\leq y_0$
    $$E^{(k)}_{test}(y)\ge \frac{(n+1)(y_0-y)}{\sum_{i=1}^ns^{(k)}(X_i,Y_i) + y_0-y} $$
so that 
$$F_{test}(y) \ge \sum_{k=1}^K \phi\left( \frac{(n+1)(y_0-y)}{\sum_{i=1}^ns^{(k)}(X_i,Y_i) + y_0-y}\right):=L(y)$$
and using bound \eqref{quantile_bound_alpha/K} :    
$$\hat Q_\alpha(y)\leq \sum_{k=1}^K \phi\left(\frac{(n+1)\hat{Q}_{\alpha/K}^k}{\sum_{i=1}^ns^{(k)}(X_i,Y_i) + y_0-y}\right):=U(y)$$
we can find $y^*$ solution of
$$U(y^*) = L(y^*).$$
Denoting $d = y_0-y^*$, and $S^k=\sum_{i=1}^ns^{(k)}(X_i,Y_i),$ we get

\begin{equation}
    \label{equality_phi}
\sum_{k=1}^K \phi\left(\frac{(n+1)\hat{Q}_{\alpha/K}^k}{S^k + d}\right) =  \sum_{k=1}^K \phi\left(\frac{(n+1)d}{S^k + d}\right)\end{equation}

which necessarily implies that $(n+1)(y_0-y^*)\leq \max_k (n+1)\hat{Q}_{\alpha/K}^k $. Otherwise, we would have $(n+1)(y_0-y^*)> (n+1)\hat{Q}_{\alpha/K}^k $ for $k=1,\dots,K$ which contradicts equality \eqref{equality_phi} since $\phi$ is strictly increasing.  This gives :

\begin{equation}
    y_{left}\ge y^* \geq y_0-\max_k\hat{Q}_{\alpha/K}^k  
\end{equation}
An analogue bound is obtained for $y_1$, where the term $(y_0-y)$ is replaced with $(y-y_1)$ in $L(y),U(y)$ for all $y\geq y_1$, which leads to :

\begin{equation}
    y_{right}\leq y_1+\max_k\hat{Q}_{\alpha/K}^k  
\end{equation}\end{proof}
\noindent Recall that $|\mathcal C_\alpha^k| = 2\hat{Q}_\alpha^k.$ 
In the event $ B_1\cup B_2\cup C$, $y_1\in \mathcal C^\Phi_\alpha$ or $y_0\in \mathcal C^\Phi_\alpha$. Therefore,
\begin{equation}
\label{bound_finalproof}
    |C^f_\alpha|\leq y_{\text{right}}-y_{\text{left}} \leq \max_k |\mathcal C_{\alpha/K}^k| + \Delta_{test}
\end{equation}
and it is obviously true in the event $A$ as we stated before.\\

Finally, to treat the case where $\phi$ is strictly decreasing, we use the fact that $$\mathcal{C}_\alpha^{ \Phi,\,\downarrow}=\mathcal{C}_\alpha^{ -\Phi,\,\uparrow}$$ Hence, the problem reduces to the previous case because $-\phi$ is non-decreasing. This concludes the proof.

\section{Experimental setup}
\label{appendix:exp_setup}
For classification, we use MNIST (70 000 samples, 784 features, 10 classes) and CIFAR-10 (60 000 samples, 3 072 features, 10 classes). The regression datasets employed in our experiments are summarized in Table \ref{tab:regression-datasets}.

\begin{table}[ht]
    \centering
    \begin{tabular}{l c r r}
        \toprule
        \textbf{OpenML ID} & \textbf{Samples (N)} & \textbf{Features (d)} \\
        \midrule
        361234 & 4\,177  & 8   \\
        361235 & 1\,503  & 5   \\
        361236 & 2\,043  & 7   \\
        361237 & 1\,030  & 8   \\
        361242 & 21\,263 & 81  \\
        361243 & 1\,059  & 116 \\
        361244 & 1\,066  & 10  \\
        361247 & 11\,934 & 14  \\
        361249 & 4\,898  & 11  \\
        \bottomrule
    \end{tabular}
    \caption{OpenML Regression datasets considered.}
    \label{tab:regression-datasets}
\end{table}

\paragraph{Regression Models.}For regression benchmarks, we used a variety of models implemented in \texttt{scikit-learn}. The feature space and target values for all regression datasets were standardized using StandardScaler, which was fit on the training data and then applied to the calibration and test sets. The models include:

\begin{itemize}
    \item \textbf{Linear Models:} We included standard LinearRegression, Lasso regression with an $L_1$ regularization penalty of $\alpha=0.1$, BayesianRidge for probabilistic linear regression, and SGDRegressor which fits a linear model using stochastic gradient descent.
    \item \textbf{Tree-based Ensembles:} We used RandomForestRegressor with 50 estimators and HistGradientBoostingRegressor. For both, the minimum leaf size was set to $\max(10, 0.001 \times n_{\text{train}})$.

    \item \textbf{Neural Network:} A MLPRegressor with two hidden layers of sizes 10 and 5, respectively, was trained using Adam optimizer with a learning rate of 0.01.
    
\end{itemize}

\paragraph{CIFAR-10 models.}We utilized several well-established deep neural network architectures implemented in PyTorch. Each model was trained for 50 epochs using Adam optimizer, a Cross-Entropy loss function, a learning rate of $1 \times 10^{-2}$, and a batch size of 64. The pool of classification models includes ResNet-56, ShuffleNetV2, VGG-16 with Batch Normalization, EfficientNet-B0, and Deep Layer Aggregation (DLA). See code for further details on the architecture.

\paragraph{MNIST Base Models.}We used four base classifiers on MNIST: a Logistic Regression with L$_2$ penalty trained via the “saga” solver (C=1.0, max\_iter=500); a Random Forest of 500 trees (max\_depth=20, max\_features=sqrt); a HistGradientBoosting classifier with 100 boosting rounds, L$_2$ regularization ($\lambda=0.1$) and early stopping; and an MLP with two hidden layers (512, 256), ReLU activations, Adam optimizer (learning rate $10^{-3}$), batch\_size=128, max\_iter=100.

\paragraph{Method and evaluation parameters.}To compute prediction‐set lengths, we evaluate on a uniform 255‐point grid spanning from $\min(y_{\mathrm{calib}})$ to $\max(y_{\mathrm{calib}})$. Quantiles are computed as exact order statistics with a time complexity of $O(n)$ time where $n$ is the calibration set size. For SACP++, we perform a parallelized search over $[-15,15]$ for regression and $[-8,8]$ for classification (to ensure numerical stability) and additionally include the $\min$ and $\max$ functions corresponding to the limit as $p\to\pm\infty$ in the candidate pool. Seeds considered are specified in the code.

\paragraph{Evaluation metrics.}We report mean and standard deviation across all 20 seeds of the following:
\begin{itemize}
  \item \textbf{Marginal coverage:} 
    \[
      \text{Coverage}
      = \frac{1}{|\mathcal{I}_{\rm test}|}\sum_{i\in\mathcal{I}_{\rm test}}
      \mathbf{1}\bigl\{Y_i\in \mathcal{C}^f_\alpha(X_i)\bigr\}. 
    \]
  \item \textbf{Average set size:}
    \[
      \text{Length}
      = \frac{1}{|\mathcal{I}_{\rm test}|}\sum_{i\in\mathcal{I}_{\rm test}}
      \bigl|\mathcal{C}^f_\alpha(X_i)\bigr|. 
    \]
\end{itemize}

\section{Additional experiments}
\label{appendix:additional_exp}
\begin{figure}[ht]
    \centering
    \includegraphics[width=\linewidth]{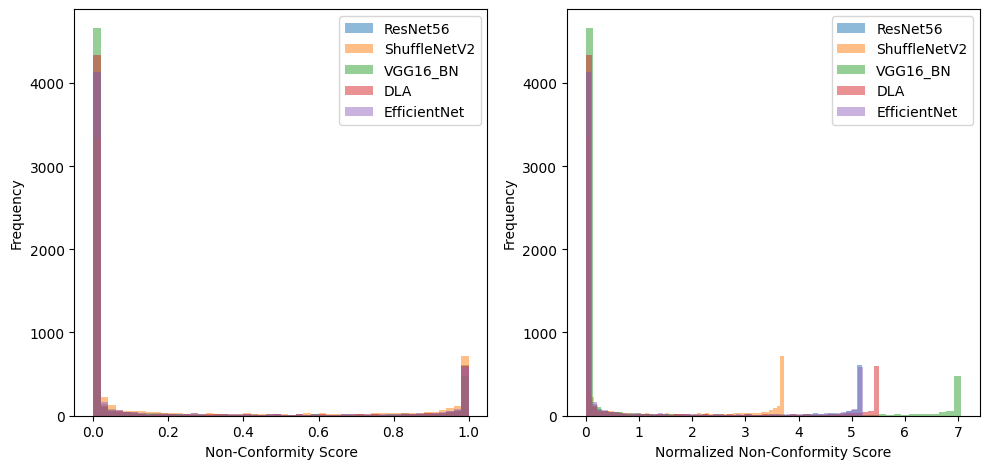}
    \caption{Distribution of raw scores (left) and e-values inspired normalization (right) for one test point from base learners on CIFAR-10, demonstrating normalization step of our SACP method. }
    \label{fig:normalization}
\end{figure}

\paragraph{On the e-values normalization step.}The normalization step is crucial for our method, as it transforms model-specific scores onto a common, comparable scale (e-values). The effect of this normalization is powerfully illustrated in our classification results on CIFAR-10 (see Fig. \ref{fig:normalization}) by the emergence of isolated spikes in the normalized distribution. These spikes correspond to the maximum NCS of 1.0. Their locations are not random; they are inversely proportional to each model's average nonconformity score on the calibration set. A spike further to the right (like VGG16\_BN's) signifies a more confident model overall. This normalization, therefore, enriches the analysis by converting a subtle difference in distribution (left plot) into a clear separation of model characteristics (right plot).

\paragraph{On the number of base models $K$.}We demonstrate the stability and competitive performance of the SACP and SACP++ methods across various datasets and numbers of base models. Regarding Coverage, SACP and SACP++ consistently achieve a mean coverage value very close to the nominal target of 0.95 for both $K=3$ (Table \ref{tab:regK3}) and $K=5$ models (Table \ref{tab:regK5}). This demonstrates the robustness and validity of our methods, in contrast to some other approaches like CR and CSA that often struggle to meet this target. When evaluating Length, SACP and SACP++ are notably effective. They frequently provide prediction intervals with significantly shorter lengths compared to other methods, such as CM and Wagg, while maintaining comparable coverage. This indicates a superior trade-off between coverage validity and interval precision. The ability of SACP++ to achieve the shortest length among methods with valid coverage on multiple datasets underscores its efficiency. This demonstrates the consistency of our methods with respect to the number of aggregated predictors, $K$.

\begin{table*}[ht]
    \scriptsize
    \centering
    \begin{tabular}{l|l|cccccc}
    Dataset (n, d) & Metric 
      & CR & CM 
      & Wagg 
      & CSA & SACP & SACP++ \\
    \hline
    \multirow{2}{*}{361234 (4177, 8)}
        & Coverage & $0.944\pm0.016$ & $0.954\pm0.011$ 
                   & $0.955\pm0.014$ 
                   & $0.945\pm0.016$ & $0.955\pm0.010$ & $0.953\pm0.008$ \\
        & Length   & $2.720\pm0.206$ & $2.877\pm0.210$ 
                   & $2.938\pm0.343$ 
                   & $2.749\pm0.314$ & $2.850\pm0.219$ & $2.781\pm0.236$ \\
    \hline
    \multirow{2}{*}{361235 (1503, 5)}
        & Coverage & $0.926\pm0.032$ & $0.969\pm0.017$ 
                   & $0.972\pm0.027$ 
                   & $0.936\pm0.036$ & $0.959\pm0.020$ & $0.950\pm0.028$ \\
        & Length   & $1.434\pm0.133$ & $1.720\pm0.115$ 
                   & $1.732\pm0.295$ 
                   & $1.456\pm0.193$ & $1.540\pm0.133$ & $1.461\pm0.115$ \\
    \hline
    \multirow{2}{*}{361236 (2043, 7)}
        & Coverage & $0.950\pm0.019$ & $0.957\pm0.018$ 
                   & $0.957\pm0.020$ 
                   & $0.931\pm0.030$ & $0.956\pm0.019$ & $0.954\pm0.021$ \\
        & Length   & $2.692\pm0.152$ & $2.755\pm0.152$ 
                   & $2.739\pm0.177$ 
                   & $2.560\pm0.193$ & $2.744\pm0.142$ & $2.717\pm0.143$ \\
    \hline
    \multirow{2}{*}{361237 (1030, 8)}
        & Coverage & $0.954\pm0.030$ & $0.975\pm0.021$ & $0.984\pm0.021$ & $0.945\pm0.041$ & $0.971\pm0.022$ & $0.966\pm0.024$ \\
        & Length   & $1.324\pm0.205$ & $1.549\pm0.199$ & $2.089\pm0.632$ & $1.244\pm0.208$ & $1.455\pm0.263$ & $1.366\pm0.213$ \\
    \hline
    \multirow{2}{*}{361242 (21263, 81)}
        & Coverage & $0.932\pm0.007$ & $0.955\pm0.005$ 
                   & $0.952\pm0.007$ 
                   & $0.952\pm0.006$ & $0.950\pm0.007$ & $0.948\pm0.007$ \\
        & Length   & $1.125\pm0.035$ & $1.243\pm0.036$ 
                   & $1.190\pm0.076$ 
                   & $1.279\pm0.058$ & $1.196\pm0.045$ & $1.165\pm0.048$ \\
    \hline
    \multirow{2}{*}{361243 (1059, 116)}
        & Coverage & $0.931\pm0.038$ & $0.963\pm0.026$ 
                   & $0.979\pm0.024$ 
                   & $0.932\pm0.046$ & $0.958\pm0.025$ & $0.950\pm0.031$ \\
        & Length   & $3.283\pm0.378$ & $3.562\pm0.392$ 
                   & $4.114\pm0.576$ 
                   & $3.277\pm0.534$ & $3.484\pm0.468$ & $3.355\pm0.398$ \\
    \hline
    \multirow{2}{*}{361244 (1066, 10)}
        & Coverage & $0.967\pm0.018$ & $0.967\pm0.018$ 
                   & $0.971\pm0.040$ 
                   & $0.946\pm0.027$ & $0.967\pm0.018$ & $0.963\pm0.019$ \\
        & Length   & $3.001\pm0.850$ & $3.003\pm0.849$ 
                   & $4.397\pm1.976$ 
                   & $1.996\pm1.042$ & $3.003\pm0.850$ & $2.997\pm0.848$ \\
    \hline
    \multirow{2}{*}{361247 (11934, 14)}
        & Coverage & $0.916\pm0.012$ & $0.968\pm0.005$ 
                   & $0.960\pm0.011$ 
                   & $0.960\pm0.011$ & $0.950\pm0.005$ & $0.949\pm0.009$ \\
        & Length   & $0.370\pm0.062$ & $0.492\pm0.073$ 
                   & $0.364\pm0.080$ 
                   & $0.408\pm0.099$ & $0.404\pm0.075$ & $0.387\pm0.062$ \\
    \hline
    \multirow{2}{*}{361249 (4898, 11)}
        & Coverage & $0.942\pm0.019$ & $0.960\pm0.015$ 
                   & $0.956\pm0.018$ 
                   & $0.950\pm0.017$ & $0.956\pm0.015$ & $0.955\pm0.013$ \\
        & Length   & $2.959\pm0.136$ & $3.155\pm0.143$ 
                   & $3.130\pm0.178$ 
                   & $3.059\pm0.147$ & $3.080\pm0.128$ & $3.017\pm0.123$ \\
    \end{tabular}
    \caption{Comparison of aggregation methods on regression datasets. $\alpha = 0.05$, $K=3$.}
    \label{tab:regK3}
\end{table*}

\begin{table*}[ht]
    \scriptsize
    \centering
    \begin{tabular}{l|l|cccccc}
    Dataset (n, d) & Metric 
      & CR & CM 
      & Wagg 
      & CSA & SACP & SACP++ \\
    \hline
    \multirow{2}{*}{361234 (4177, 8)}
        & Coverage & $0.940\pm0.012$ & $0.951\pm0.013$ 
                   & $0.949\pm0.009$ 
                   & $0.939\pm0.008$ & $0.947\pm0.013$ & $0.944\pm0.010$ \\
        & Length   & $2.713\pm0.144$ & $2.978\pm0.180$ 
                   & $2.964\pm0.388$ 
                   & $2.740\pm0.173$ & $2.825\pm0.219$ & $2.740\pm0.201$ \\
    \hline
    \multirow{2}{*}{361235 (1503, 5)}
        & Coverage & $0.917\pm0.024$ & $0.954\pm0.023$ 
                   & $0.951\pm0.019$ 
                   & $0.917\pm0.023$ & $0.949\pm0.028$ & $0.940\pm0.035$ \\
        & Length   & $1.954\pm0.081$ & $2.724\pm0.135$ 
                   & $1.797\pm0.091$ 
                   & $1.954\pm0.081$ & $2.138\pm0.420$ & $1.844\pm0.439$ \\
    \hline
    \multirow{2}{*}{361236 (2043, 7)}
        & Coverage & $0.931\pm0.026$ & $0.946\pm0.024$ 
                   & $0.952\pm0.023$ 
                   & $0.920\pm0.031$ & $0.949\pm0.024$ & $0.944\pm0.027$ \\
        & Length   & $2.590\pm0.164$ & $2.705\pm0.149$ 
                   & $2.747\pm0.145$ 
                   & $2.479\pm0.168$ & $2.720\pm0.153$ & $2.661\pm0.159$ \\
    \hline
    \multirow{2}{*}{361237 (1030, 8)}
        & Coverage & $0.977\pm0.012$ & $0.980\pm0.011$ & $0.976\pm0.013$ & $0.975\pm0.014$ & $0.978\pm0.012$ & $0.976\pm0.013$ \\
        & Length   & $1.705\pm0.035$ & $1.840\pm0.040$ & $1.720\pm0.036$ & $1.695\pm0.038$ & $1.715\pm0.034$ & $1.700\pm0.037$ \\
    \hline
    \multirow{2}{*}{361242 (21263, 81)}
        & Coverage & $0.921\pm0.006$ & $0.965\pm0.004$ 
                   & $0.950\pm0.004$ 
                   & $0.948\pm0.004$ & $0.950\pm0.003$ & $0.950\pm0.004$ \\
        & Length   & $1.260\pm0.028$ & $1.657\pm0.015$ 
                   & $1.186\pm0.064$ 
                   & $1.432\pm0.058$ & $1.325\pm0.026$ & $1.229\pm0.030$ \\
    \hline
    \multirow{2}{*}{361243 (1059, 116)}
        & Coverage & $0.940\pm0.031$ & $0.962\pm0.022$ 
                   & $0.970\pm0.023$ 
                   & $0.936\pm0.038$ & $0.949\pm0.028$ & $0.940\pm0.035$ \\
        & Length   & $3.348\pm0.455$ & $3.594\pm0.428$ 
                   & $3.951\pm0.598$ 
                   & $3.501\pm0.512$ & $3.489\pm0.420$ & $3.387\pm0.439$ \\
    \hline
    \multirow{2}{*}{361244 (1066, 10)}
        & Coverage & $0.968\pm0.017$ & $0.968\pm0.017$ 
                   & $0.946\pm0.070$ 
                   & $0.938\pm0.033$ & $0.968\pm0.017$ & $0.966\pm0.017$ \\
        & Length   & $2.925\pm1.019$ & $2.944\pm1.018$ 
                   & $4.602\pm3.092$ 
                   & $1.948\pm1.023$ & $2.943\pm1.021$ & $2.926\pm1.021$ \\
    \hline
    \multirow{2}{*}{361247 (11934, 14)}
        & Coverage & $0.900\pm0.007$ & $0.980\pm0.003$ 
                   & $0.968\pm0.004$ 
                   & $0.943\pm0.017$ & $0.950\pm0.002$ & $0.948\pm0.007$ \\
        & Length   & $0.596\pm0.047$ & $1.517\pm0.024$ 
                   & $0.445\pm0.077$ 
                   & $0.563\pm0.093$ & $0.481\pm0.070$ & $0.408\pm0.044$ \\
    \hline
    \multirow{2}{*}{361249 (4898, 11)}
        & Coverage & $0.943\pm0.018$ & $0.964\pm0.014$ 
                   & $0.961\pm0.021$ 
                   & $0.952\pm0.013$ & $0.963\pm0.015$ & $0.958\pm0.011$ \\
        & Length   & $3.205\pm0.165$ & $3.591\pm0.179$ 
                   & $3.220\pm0.186$ 
                   & $3.228\pm0.081$ & $3.407\pm0.162$ & $3.246\pm0.110$ \\
    \end{tabular}
    \caption{Comparison of aggregation methods on regression datasets. $\alpha = 0.05$, $K=5$.}
    \label{tab:regK5}
\end{table*}

\paragraph{On the $\alpha$–stability.}SACP++ consistently achieves nominal coverage and yields the shortest prediction set sizes across all values of $\alpha$ considered on both CIFAR-10 (Table~\ref{tab:cifar10-multialpha}) and MNIST (Table~\ref{tab:mnist-multialpha}) with only a single exception, thereby demonstrating superior efficiency and $\alpha$-stability.

\begin{table*}[ht]
  \scriptsize
  \centering
  \setlength{\tabcolsep}{2pt}
  \begin{tabular}{lc|c|c|c||c|c|c|c}
    \toprule
    \textbf{Method} & \multicolumn{4}{c}{\textbf{Coverage}} & \multicolumn{4}{c}{\textbf{Length}} \\
                    & $\alpha=0.025$ & $\alpha=0.05$ & $\alpha=0.075$ & $\alpha=0.1$ & 
                      $\alpha=0.025$ & $\alpha=0.05$ & $\alpha=0.075$ & $\alpha=0.1$ \\
    \midrule
    \multicolumn{9}{c}{\textit{Base Models}} \\
    \midrule
    ResNet56        & 0.974 $\pm$ 0.003 & 0.949 $\pm$ 0.002 & 0.925 $\pm$ 0.003 & 0.901 $\pm$ 0.004 & 
                      2.411 $\pm$ 0.091 & 1.833 $\pm$ 0.062 & 1.540 $\pm$ 0.045 & 1.352 $\pm$ 0.040 \\
    ShuffleNetV2    & 0.974 $\pm$ 0.003 & 0.949 $\pm$ 0.003 & 0.924 $\pm$ 0.003 & 0.899 $\pm$ 0.004 & 
                      3.030 $\pm$ 0.117 & 2.322 $\pm$ 0.089 & 1.936 $\pm$ 0.065 & 1.686 $\pm$ 0.056 \\
    VGG16\_BN       & 0.975 $\pm$ 0.003 & 0.951 $\pm$ 0.003 & 0.924 $\pm$ 0.003 & 0.899 $\pm$ 0.004 & 
                      2.130 $\pm$ 0.130 & 1.531 $\pm$ 0.068 & 1.272 $\pm$ 0.038 & 1.130 $\pm$ 0.025 \\
    DLA             & 0.975 $\pm$ 0.003 & 0.950 $\pm$ 0.003 & 0.925 $\pm$ 0.004 & 0.899 $\pm$ 0.005 & 
                      2.122 $\pm$ 0.118 & 1.620 $\pm$ 0.063 & 1.375 $\pm$ 0.046 & 1.218 $\pm$ 0.037 \\
    EfficientNet    & 0.975 $\pm$ 0.003 & 0.949 $\pm$ 0.004 & 0.924 $\pm$ 0.004 & 0.900 $\pm$ 0.005 & 
                      2.723 $\pm$ 0.110 & 1.929 $\pm$ 0.067 & 1.573 $\pm$ 0.047 & 1.366 $\pm$ 0.033 \\
    \midrule
    \multicolumn{9}{c}{\textit{Aggregation Methods}} \\
    \midrule
    BL       & 0.975 $\pm$ 0.003 & 0.951 $\pm$ 0.003 & 0.924 $\pm$ 0.003 &   0.899 $\pm$ 0.004 & 
                      2.130 $\pm$ 0.130 & 1.531 $\pm$ 0.068 & 1.272 $\pm$ 0.038 & 1.130 $\pm$ 0.025 \\
    CR              & 0.969 $\pm$ 0.001 & 0.948 $\pm$ 0.005 & 0.931 $\pm$ 0.004 & 0.908 $\pm$ 0.006 & 
                      1.552 $\pm$ 0.042 & 1.506 $\pm$ 0.054 & 1.038 $\pm$ 0.015 & 1.194 $\pm$ 0.029 \\
    CM              & 0.999 $\pm$ 0.001 & 0.988 $\pm$ 0.002 & 0.951 $\pm$ 0.002 & 0.971 $\pm$ 0.003 & 
                      2.155 $\pm$ 0.063 & 1.33 $\pm$ 0.026 & 1.322 $\pm$ 0.022 & 1.570 $\pm$ 0.050 \\
    Wagg            & 0.974 $\pm$ 0.003 & 0.948 $\pm$ 0.003 & 0.923 $\pm$ 0.004 & 0.898 $\pm$ 0.005 & 
                      1.658 $\pm$ 0.044 & 1.292 $\pm$ 0.028 & 1.128 $\pm$ 0.019 & 1.032 $\pm$ 0.014 \\
    CSA             & 0.976 $\pm$ 0.004 & 0.950 $\pm$ 0.005 & 0.924 $\pm$ 0.005 & 0.898 $\pm$ 0.007 & 
                      1.675 $\pm$ 0.079 & 1.294 $\pm$ 0.037 & 1.134 $\pm$ 0.023 & 1.038 $\pm$ 0.018 \\
    SACP            & 0.975 $\pm$ 0.003 & 0.950 $\pm$ 0.003 & 0.925 $\pm$ 0.004 & 0.899 $\pm$ 0.004 & 
                      1.746 $\pm$ 0.043 & 1.308 $\pm$ 0.033 & 1.132 $\pm$ 0.019 & 1.034 $\pm$ 0.014 \\
    SACP++          & 0.975 $\pm$ 0.003 & 0.949 $\pm$ 0.003 & 0.923 $\pm$ 0.004 & 0.898 $\pm$ 0.004 & 
                      \textbf{1.629 $\pm$ 0.051} & \textbf{1.281 $\pm$ 0.027} & \textbf{1.118 $\pm$ 0.016} & \textbf{1.028 $\pm$ 0.011} \\
    \bottomrule
  \end{tabular}
  \caption{Performance on CIFAR-10 across different $\alpha$ levels.}
  \label{tab:cifar10-multialpha}
\end{table*}

\begin{table*}[ht]
  \scriptsize
  \centering
  \setlength{\tabcolsep}{2pt}
  \begin{tabular}{lc|c|c|c||c|c|c|c}
    \toprule
    \multirow{2}{*}{\textbf{Method}} 
      & \multicolumn{4}{c}{\textbf{Coverage}} 
      & \multicolumn{4}{c}{\textbf{Length}} \\
    & $\alpha=0.025$ & $\alpha=0.05$ & $\alpha=0.075$ & $\alpha=0.1$
    & $\alpha=0.025$ & $\alpha=0.05$ & $\alpha=0.075$ & $\alpha=0.1$ \\
    \midrule
    \multicolumn{9}{c}{\textit{Base Models}} \\
    \midrule
    Logistic     & 0.976 $\pm$ 0.002 & 0.951 $\pm$ 0.004 & 0.926 $\pm$ 0.005 & 0.902 $\pm$ 0.005
                 & 1.394 $\pm$ 0.034 & 1.109 $\pm$ 0.016 & 1.012 $\pm$ 0.008 & 0.956 $\pm$ 0.007 \\
    RandomForest & 0.976 $\pm$ 0.003 & 0.951 $\pm$ 0.005 & 0.926 $\pm$ 0.006 & 0.901 $\pm$ 0.006
                 & 1.027 $\pm$ 0.007 & 0.970 $\pm$ 0.005 & 0.936 $\pm$ 0.006 & 0.908 $\pm$ 0.006 \\
    HistGB       & 0.975 $\pm$ 0.003 & 0.950 $\pm$ 0.005 & 0.925 $\pm$ 0.005 & 0.901 $\pm$ 0.006
                 & 0.996 $\pm$ 0.005 & 0.958 $\pm$ 0.005 & 0.929 $\pm$ 0.005 & 0.903 $\pm$ 0.005 \\
    MLP          & 0.975 $\pm$ 0.002 & 0.950 $\pm$ 0.003 & 0.927 $\pm$ 0.004 & 0.908 $\pm$ 0.010
                 & 0.997 $\pm$ 0.004 & 0.958 $\pm$ 0.004 & 0.932 $\pm$ 0.005 & 0.912 $\pm$ 0.010 \\
    \midrule
    \multicolumn{9}{c}{\textit{Aggregation Methods}} \\
    \midrule
    BL    & 0.975 $\pm$ 0.003 & 0.950 $\pm$ 0.005 & 0.925 $\pm$ 0.005 & 0.901 $\pm$ 0.006
          & 0.996 $\pm$ 0.005 & 0.958 $\pm$ 0.004 & 0.929 $\pm$ 0.005 & 0.903 $\pm$ 0.005 \\
    CR    & 0.980 $\pm$ 0.003 & 0.962 $\pm$ 0.003 & 0.944 $\pm$ 0.004 & 0.926 $\pm$ 0.005
          & 1.013 $\pm$ 0.006 & 0.975 $\pm$ 0.004 & 0.952 $\pm$ 0.005 & 0.931 $\pm$ 0.005 \\
    CM    & 0.994 $\pm$ 0.001 & 0.961 $\pm$ 0.018 & 0.977 $\pm$ 0.002 & 0.967 $\pm$ 0.003
          & 1.125 $\pm$ 0.017 & 0.975 $\pm$ 0.024 & 1.009 $\pm$ 0.005 & 0.988 $\pm$ 0.004 \\
    Wagg  & 0.974 $\pm$ 0.002 & 0.950 $\pm$ 0.004 & 0.925 $\pm$ 0.005 & 0.901 $\pm$ 0.005
          & \textbf{0.986 $\pm$ 0.004} & 0.955 $\pm$ 0.004 & 0.928 $\pm$ 0.005 & 0.903 $\pm$ 0.005 \\
    CSA   & 0.974 $\pm$ 0.003 & 0.952 $\pm$ 0.004 & 0.926 $\pm$ 0.005 & 0.900 $\pm$ 0.006
          & 0.990 $\pm$ 0.005 & 0.958 $\pm$ 0.005 & 0.930 $\pm$ 0.005 & 0.903 $\pm$ 0.006 \\
    SACP  & 0.975 $\pm$ 0.002 & 0.951 $\pm$ 0.004 & 0.926 $\pm$ 0.004 & 0.901 $\pm$ 0.005
          & 0.991 $\pm$ 0.003 & 0.956 $\pm$ 0.004 & 0.929 $\pm$ 0.004 & 0.904 $\pm$ 0.005 \\
    SACP++& 0.974 $\pm$ 0.002 & 0.948 $\pm$ 0.004 & 0.923 $\pm$ 0.004 & 0.897 $\pm$ 0.005
          & 0.988 $\pm$ 0.003 & \textbf{0.954 $\pm$ 0.004} & \textbf{0.927 $\pm$ 0.004} & \textbf{0.900 $\pm$ 0.006} \\
    \bottomrule
  \end{tabular}
  \caption{Performance on MNIST across different $\alpha$ levels.}
  \label{tab:mnist-multialpha}
\end{table*}

\end{document}